\documentclass[runningheads]{llncs}
\usepackage[T1]{fontenc}
\usepackage{graphicx}
\usepackage{comment}
\usepackage{amsmath,amssymb}
\usepackage{color}
\usepackage{url}
\usepackage{hyperref}
\usepackage{subcaption}
\usepackage{booktabs}
\usepackage{cuted}

\usepackage[capitalize]{cleveref}
\crefname{section}{Sec.}{Secs.}
\Crefname{section}{Section}{Sections}
\Crefname{table}{Table}{Tables}
\crefname{table}{Tab.}{Tabs.}

\begin{document}
\title{Evidential Deep Learning for Class-Incremental Semantic Segmentation}%

\author{Karl Holmquist \inst{1} \and
	Lena Klas\'en\inst{1,2} \and
	Michael Felsberg\inst{1}}
	
\authorrunning{K. Holmquist et al.}

\institute{Linköping University, Sweden 
\and Sweden Office of the National Police Commissioner, The Swedish Police Authority\\
\email{}}

\maketitle              %
\begin{abstract}
    Class-Incremental Learning is a challenging problem in machine learning that aims to extend previously trained neural networks with new classes. 
    This is especially useful if the system is able to classify new objects despite the original training data being unavailable.
    While the semantic segmentation problem has received less attention than classification, it poses distinct problems and challenges since previous and future target classes can be unlabeled in the images of a single increment. In this case, the background, past and future classes are correlated and there exist a \textit{background-shift}. %
    
    In this paper, we address the problem of how to model unlabeled classes while avoiding spurious feature clustering of future uncorrelated classes. 
    We propose to use Evidential Deep Learning to model the evidence of the classes as a Dirichlet distribution. 
    Our method factorizes the problem into a separate foreground class probability, calculated by the expected value of the Dirichlet distribution, and an unknown class (background) probability corresponding to the uncertainty of the estimate.
    In our novel formulation, the background probability is implicitly modeled, avoiding the feature space clustering
    that comes from forcing the model to output a high background score for pixels that are not labeled as objects.
    Experiments on the incremental Pascal VOC, and ADE20k benchmarks show that our method is superior to state-of-the-art, especially when repeatedly learning new classes with increasing number of increments.
    
\end{abstract}

\section{Introduction}
Semantic segmentation is a challenging fundamental problem in computer vision with applications to many real-world tasks which require detailed knowledge regarding the surrounding environment. While the introduction of new architectures e.g., Convolutional Neural Networks (CNNs) \cite{chen2017deeplab,krizhevsky2012imagenet} and transformers \cite{liu2021swin,zheng2021rethinking}, as well as large-scale annotated datasets \cite{pascal-voc-2012,zhou2019semantic}, has led to significant improvements in semantic segmentation, the models are typically constrained by pre-defined sets of classes.
Thus, if current semantic segmentation methods are to be extended to new classes, retraining of the entire network and availability of both the old and the new training data, fully annotated with all classes, is required.

Instead, Class-incremental Semantic Segmentation (CISS) \cite{castro2018end,he2021tale,zhang2006class} aims at expanding the set of known classes of a trained model by continual learning, addressing the problem of adding new classes to an existing network while avoiding \textit{catastrophic forgetting} that causes the performance on the old classes to degrade.

\begin{figure}[b!]
    \centering
    \includegraphics[width=\linewidth]{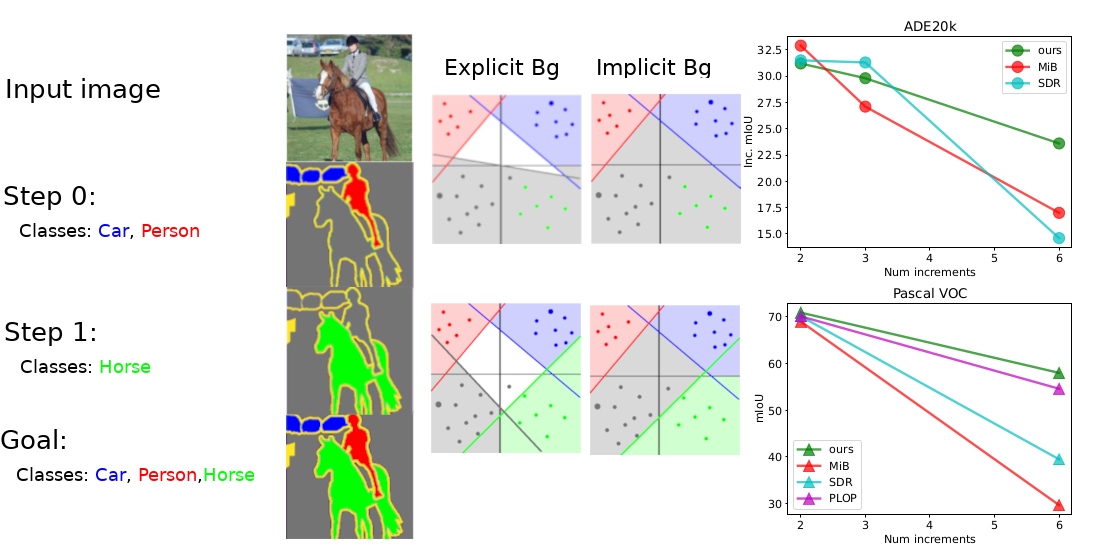}
    \caption{The leftmost part of the figure illustrates the task, which is to incrementally learn all of the classes, \textit{car}, \textit{person}, \textit{horse}, and the \textit{background}. In the first step, the training data contains labels of the first two classes and in the second step only of the third class with the previous two marked as background. 
    As illustrated in the middle using a linear classifier, an explicit background model requires a large change to separate the previous background class into the two subclasses, \textit{horse} and \textit{background.}.
    Contrary to our proposed implicit background model which classifies uncertain parts of the feature space as background, an explicit background model instead interpolate the class from the closest linear classifier.
    The right-most part of the image shows the performance of the current state-of-the-art models based on the number of increments in each task. Our method is clearly more robust to the increasing number of incremental steps in the learning.
    }
    \label{fig:first}
\end{figure}

The topic of continual learning has been studied in various fields, such as image classification \cite{aljundi2018memory,kirkpatrick2017overcoming,liu2020mnemonics}, object detection \cite{liu2020continual}, robotics \cite{lesort2019continual}, and long-term visual tracking \cite{LT-vot}.
More recently, continual learning has also been applied to semantic segmentation in the form of class-incremental semantic segmentation \cite{Cermelli_2020_CVPR,douillard2021plop,Maracani_2021_ICCV,Michieli_2021_CVPR}.

While standard semantic segmentation operates under a closed-world assumption where all possible classes are known during training. CSS is based on an open-world assumption \cite{drummond2006open}, in which the background class is a mixture of different unlabeled classes, both previously learned and future classes, and a generic background class.
The open-world setting makes the common explicit background modeling approach less suitable since the weight layer of the background class requires large changes between the increments to handle the \textit{background shift}, see \cref{fig:first}.

Previous methods have addressed this problem using in-painting techniques \cite{douillard2021plop} and by modeling the background as a mixture of a background class and multiple unlabeled classes \cite{Cermelli_2020_CVPR}. However, both of these approaches are based on explicit background modeling, which clusters future classes together into the background class.

Contrary to previous works, our approach avoids this confusion of future classes and background by implicitly modeling the background as the absence of evidence for any of the known labeled classes, facilitating repeatedly learning new classes without the degradation of results seen in current state-of-the-art methods.

\paragraph{Our main contributions} are:

\begin{itemize}
    \item We introduce a novel approach to implicitly model the background based on Evidential Deep Learning for Class-incremental semantic segmentation.
    \item We perform multiple ablation studies to validate our approach, which gives evidence about its soundness.
    \item We perform exhaustive large-scale experiments and show that our method is superior to the state-of-the-art on class-incremental Pascal VOC and ADE20k, especially for a larger number of increments.
\end{itemize}

\section{Related work}

\paragraph{Continual Learning:}
The most prevailing type of approach to deep learning, especially for vision-based segmentation and classification, is done in a batched fashion, where all data and labels are provided before the learning starts. Traditionally, this has prohibited incrementally adding new knowledge to an already trained model since the previous knowledge tends to be easily forgotten and cause catastrophic forgetting.

Class-incremental learning was first introduced for classification where new classes are added to the set of known classes of an image classification model \cite{li2017learning}. Since then, multiple approaches to this type of continual learning have been proposed.
Most of these methods employs one or multiple of the following approaches: regularization-based approaches \cite{aljundi2018memory,kirkpatrick2017overcoming}, topology-preserving approaches \cite{tao2020few,tao2020topology,Yu_2020_CVPR}, or rehearsal-based approaches \cite{Mi_2020_CVPR_Workshops,rebuffi2017icarl}. 
Regularization-based approaches can be separated into distillation-based methods and penalty-computing methods \cite{Michieli_2021_CVPR}. 

Distillation-based methods \cite{hinton2015distilling,li2017learning,Zhang_2020_WACV,Zhao_2020_CVPR} use knowledge distillation to transfer knowledge from one or multiple \textit{teachers} trained on previous tasks \cite{choi2021dual} to the new \textit{student} network. 
Knowledge distillation is commonly used in most approaches above and can be applied both on the output level and the internal feature level \cite{douillard2021plop} often in the form of a cross-entropy or $\ell_2$ loss.

Penalty-computing methods aim at directing the parameter updates so that the new knowledge is interfering as little as possible with previous knowledge \cite{aljundi2018memory,kirkpatrick2017overcoming} and topology-preserving methods have a similar aim to feature-level distillation-based approaches in that they preserve the feature-space topology \cite{tao2020few,tao2020topology}.

The other common category is formed by rehearsal-based approaches, which either maintain a small sample set of previous tasks \cite{Mi_2020_CVPR_Workshops,rebuffi2017icarl} or use a Generative Adversarial Network (GAN) for generating samples of previous tasks  \cite{liu2020generative,wu2018incremental,xiang2019incremental}.

\paragraph{Class-Incremental Semantic Segmentation:}
Recently, the class-incremental problem was formulated for semantic segmentation \cite{Cermelli_2020_CVPR,Michieli_2021_CVPR,michieli2019incremental}, which not only encounters the issue of catastrophic forgetting, but also \textit{background shift} \cite{douillard2021plop}.
The background shift arises due to the inconsistent interpretation of background between separate increments, which contains also the respectively unlabeled classes.

One of the first methods to address this issue together with catastrophic forgetting was proposed by Cermelli et al. \cite{Cermelli_2020_CVPR} who modeled the background as a mixture of the background class and unlabeled classes. However, this explicit background model constrains future classes to be similar in feature space. We argue that implicit modeling of the background class is the most appropriate formulation to address an open-world assumption on the class sets, enabling both previous and future classes to be part of the current background class.

Unlabeled previously seen classes was partly addressed by the pseudo-labeling in PLOP \cite{douillard2021plop}, who used an uncertainty measure based on median entropy to decide if the model was confident enough to use the predicted label or not. In addition to this, they also expanded the knowledge-distillation term used by \cite{Cermelli_2020_CVPR} to a multi-scale local distillation.

Another aspect of importance to allow continual learning is the underlying learned feature space.
SDR, based on representation learning \cite{Michieli_2021_CVPR} applied a clustering approach and a sparsity loss to improve the feature space and thus facilitate continuous learning.

In addition to the approaches above, 
RECALL \cite{Maracani_2021_ICCV} applies either a web crawler or a GAN to expand the training data with examples of previous classes and annotate these using a pseudo labeling approach based on the previous model. This significantly improved the performance when using multiple increments and a few classes per increment.

Our method is independent of the network architecture and maintains a simplistic distillation loss while it does not use any extra data as in \cite{Maracani_2021_ICCV}. However, expanding our method with these components would be straightforward.

\paragraph{Evidential Deep Learning (EDL)}
was proposed by Sensoy et.al. \cite{sensoy2018evidential} as a more computationally efficient method compared to Bayesian inference and ensemble methods for estimating the epistemic uncertainty, i.e. the uncertainty of the predictions of a neural network. Instead of training multiple models and estimating the variance in their predictions, a single model is trained to output the parameters of a probability distribution, the normal distribution for regression problems, and the Dirichlet distribution for classification.
The use of EDL has been shown to be useful for open-world action recognition settings \cite{Evidential_actions_2021_ICPR}.

EDL for classification problems is based on subjective logic, which assigns a belief, $b_i$, to each singleton (each class in our case) of a \textit{frame of discernment}, i.e. the set of possible classes. These beliefs correspond to an observer's belief of each of the classes being true. However, the observer is normally not completely confident in their belief, which leads to an additional amount of uncertainty, $u$. The total belief and uncertainty mass is defined to always be one, i.e.
\begin{equation}
    u + \sum_i b_i = 1 .
\end{equation}

To model these beliefs, they are formulated in terms of the concentration parameters of a Dirichlet distribution, which PDF lies on an $N$d-simplex.
The model is trained to predict scores, $z_i$, for each of the classes. These scores are rectified, normally by a ReLU or an exponential function, and offset to calculate the concentration parameters of a Dirichlet distribution,
\begin{equation}
    \alpha_i = \text{Rectifier}(z_i) + 1 = e_i + 1 .
\end{equation}

To reconnect with the subjective logic heritage, the belief mass and the uncertainty mass are defined in terms of the evidence, $e_i$, as
\begin{equation}
    b_i = \frac{e_i}{\sum_k (e_k + 1)}, \: u = \frac{K}{\sum_k (e_k + 1)} .
    \label{eq:belief}
\end{equation}

The class probabilities are obtained from the Dirichlet distribution as,
\begin{equation}
    p_i^\text{fg} = \frac{\alpha_i}{\sum_k \alpha_k} .
    \label{eq:prob}
\end{equation}

In this paper, we adapt the Evidential Deep Learning framework to the continual semantic segmentation problem and formulate the training to account for the incremental nature, which to the best of our knowledge, has not been done before.

\begin{figure*}[b!]
    \centering
    \subfloat[\label{fig:inferred:evf_rgb} RGB image]{
    \includegraphics[width=0.19\linewidth]{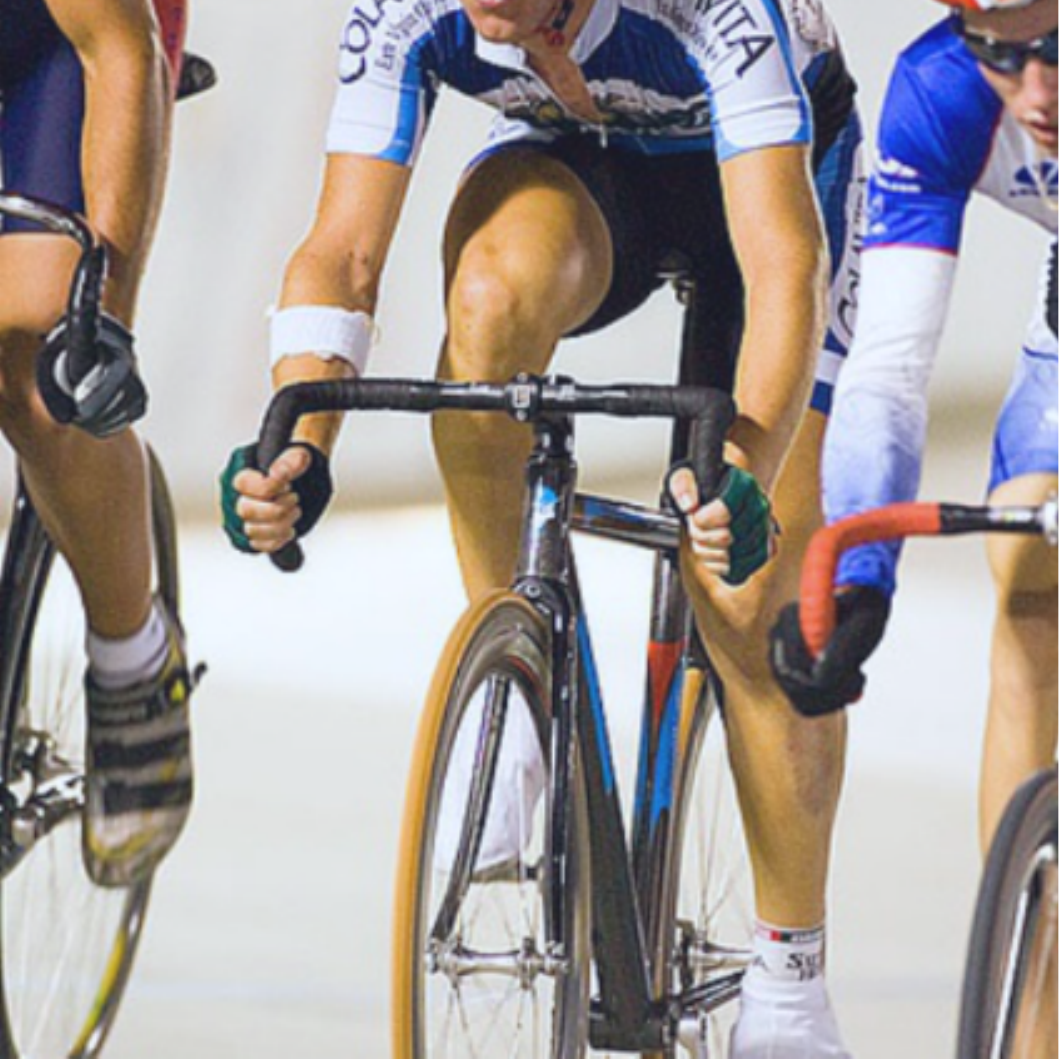}
    }
    \subfloat[\label{fig:inferred:evf_uncert_bkg} $P_\text{Bkg} =u$]{
    \includegraphics[width=0.19\linewidth]{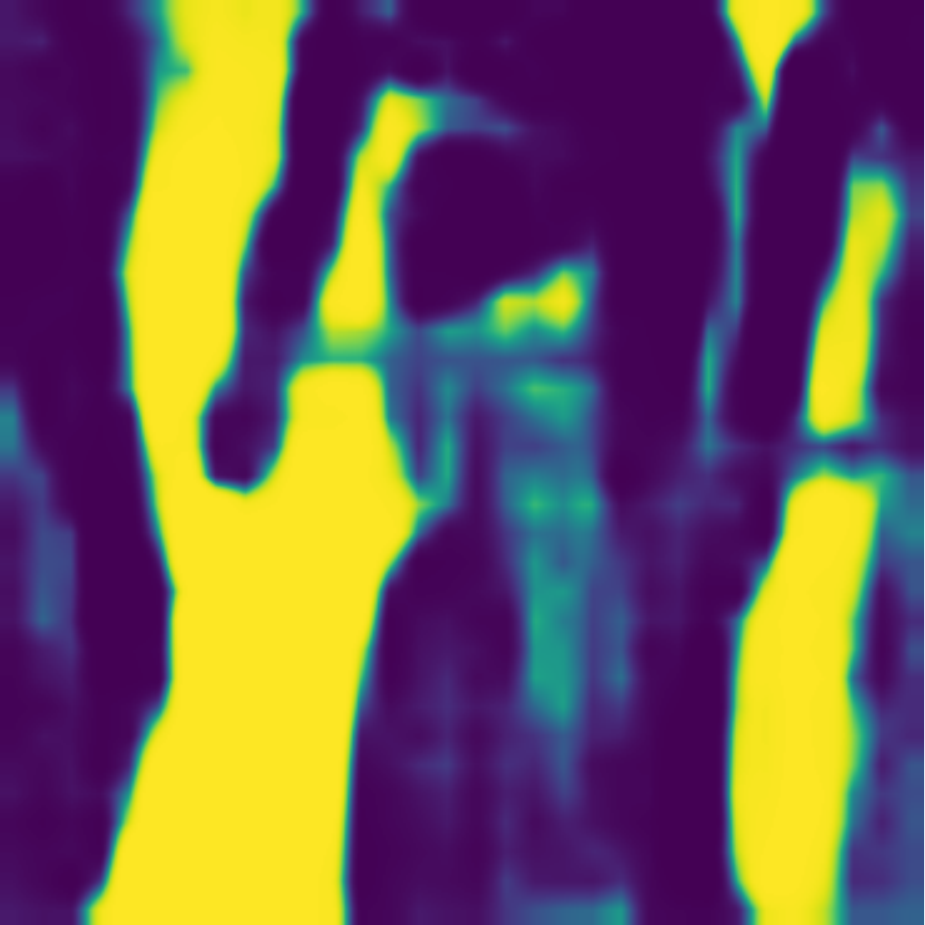}
    }
    \subfloat[\label{fig:inferred:evf_uncert_bike} $P_{\text{Bicycle}}$]{
    \includegraphics[width=0.19\linewidth]{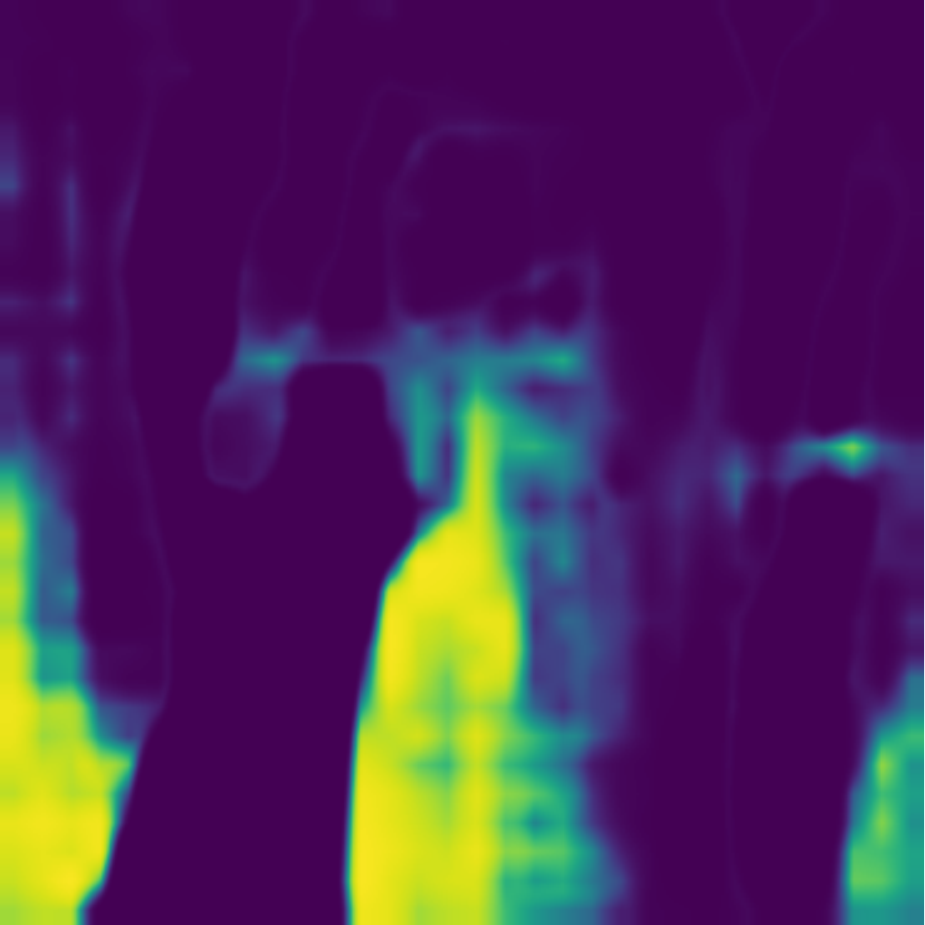}
    }
    \subfloat[\label{fig:inferred:evf_uncert_person} $P_{\text{Person}}$]{
    \includegraphics[width=0.19\linewidth]{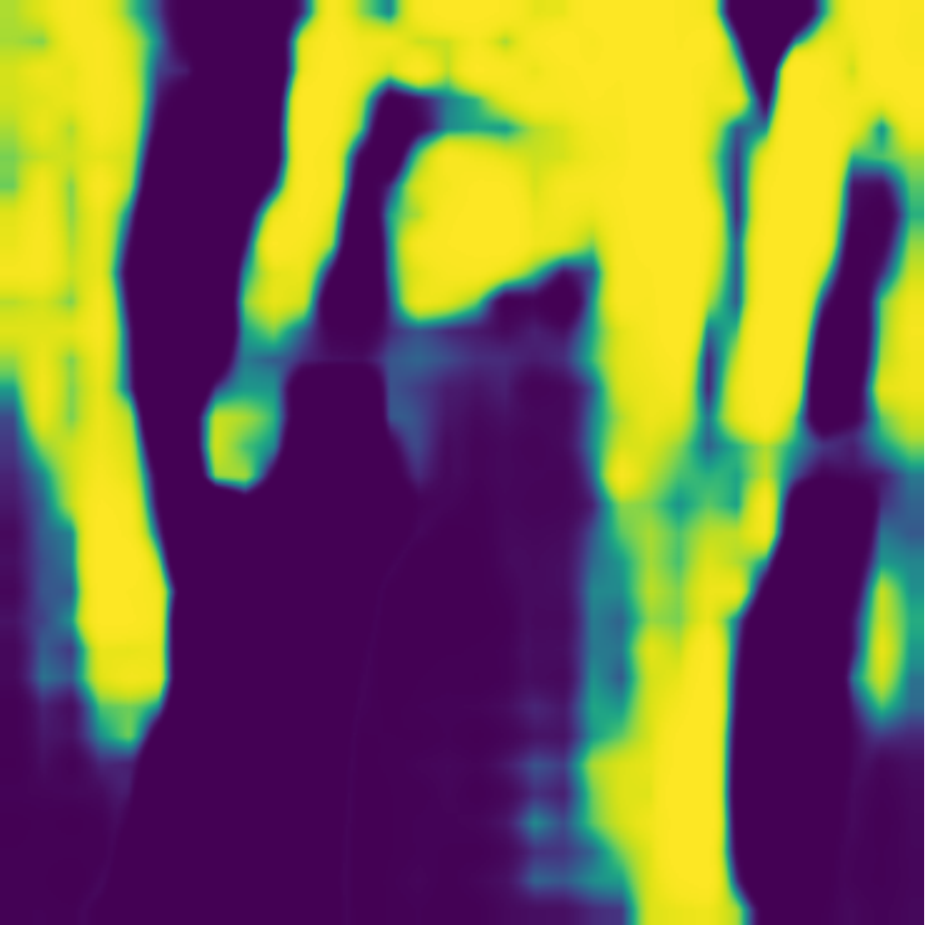}
    }
    \subfloat[\label{fig:inferred:evf_pred} Pred.]{
    \includegraphics[width=0.19\linewidth]{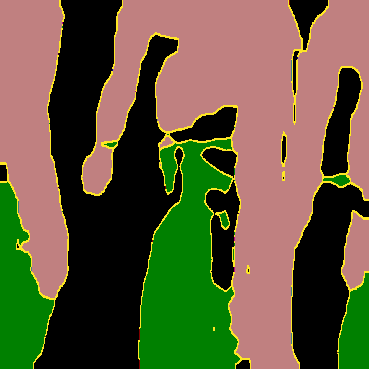}
    }\\
    \subfloat[\label{fig:inferred:gt} GT labels]{
    \includegraphics[width=0.19\linewidth]{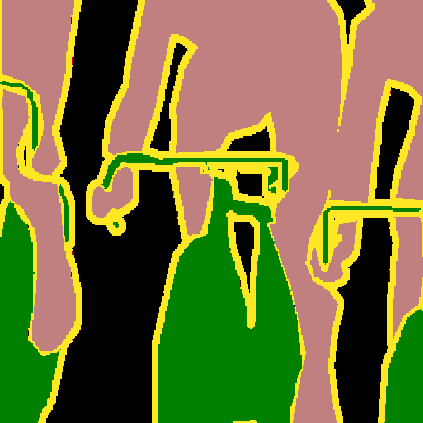}
    }
    \subfloat[\label{fig:inferred:mib_uncert_bkg} MiB - $P_\text{Bkg}$]{
    \includegraphics[width=0.19\linewidth]{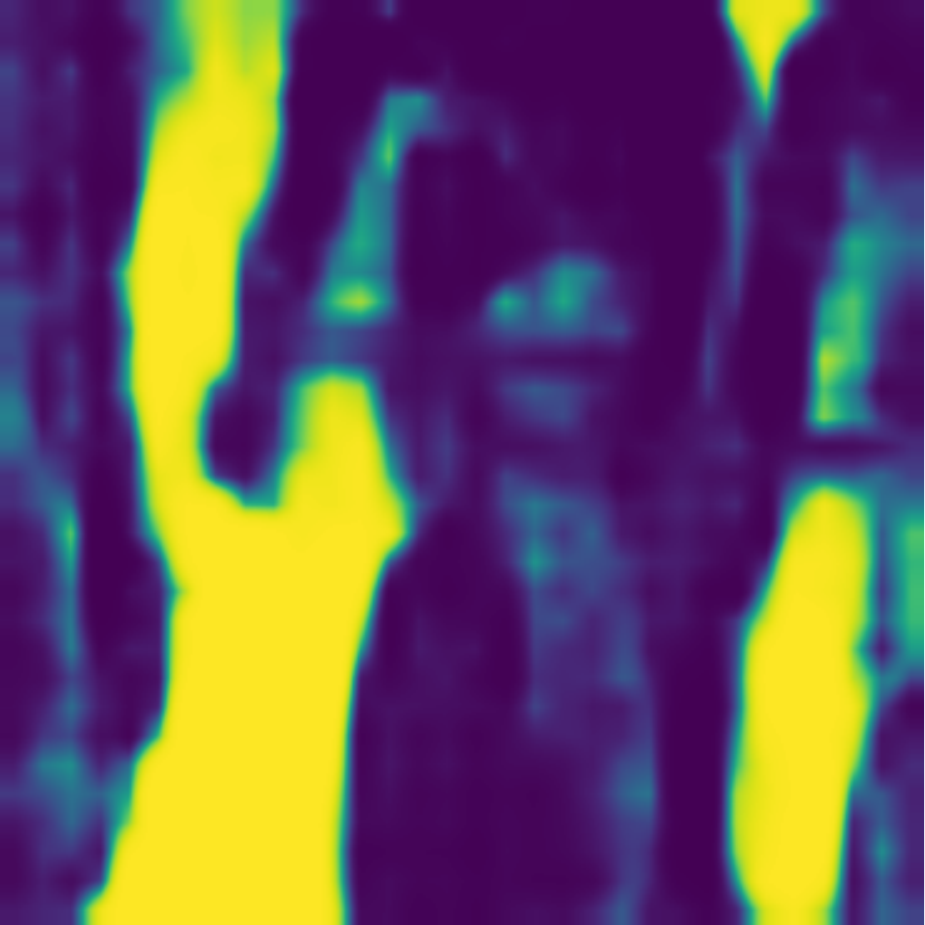}
    }
    \subfloat[\label{fig:inferred:mib_uncert_bike} MiB - $P_{\text{Bicycle}}$]{
    \includegraphics[width=0.19\linewidth]{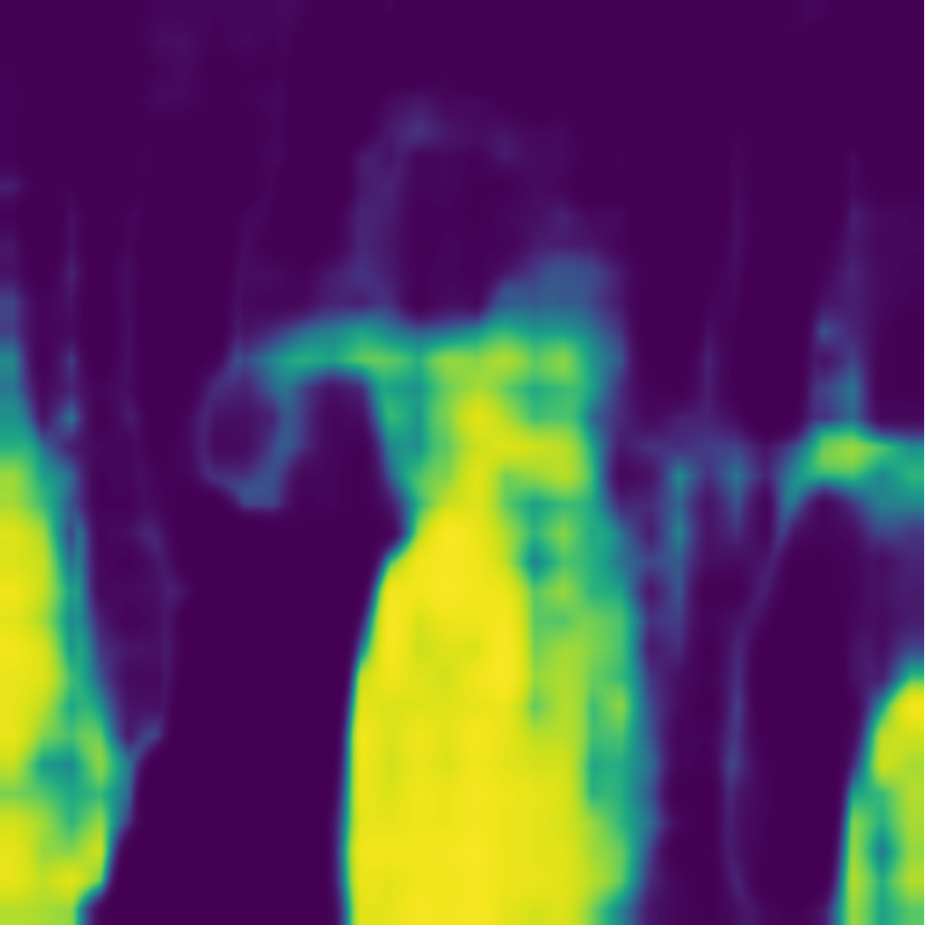}
    }
    \subfloat[\label{fig:inferred:mib_uncert_person} MiB - $P_{\text{Person}}$]{
    \includegraphics[width=0.19\linewidth]{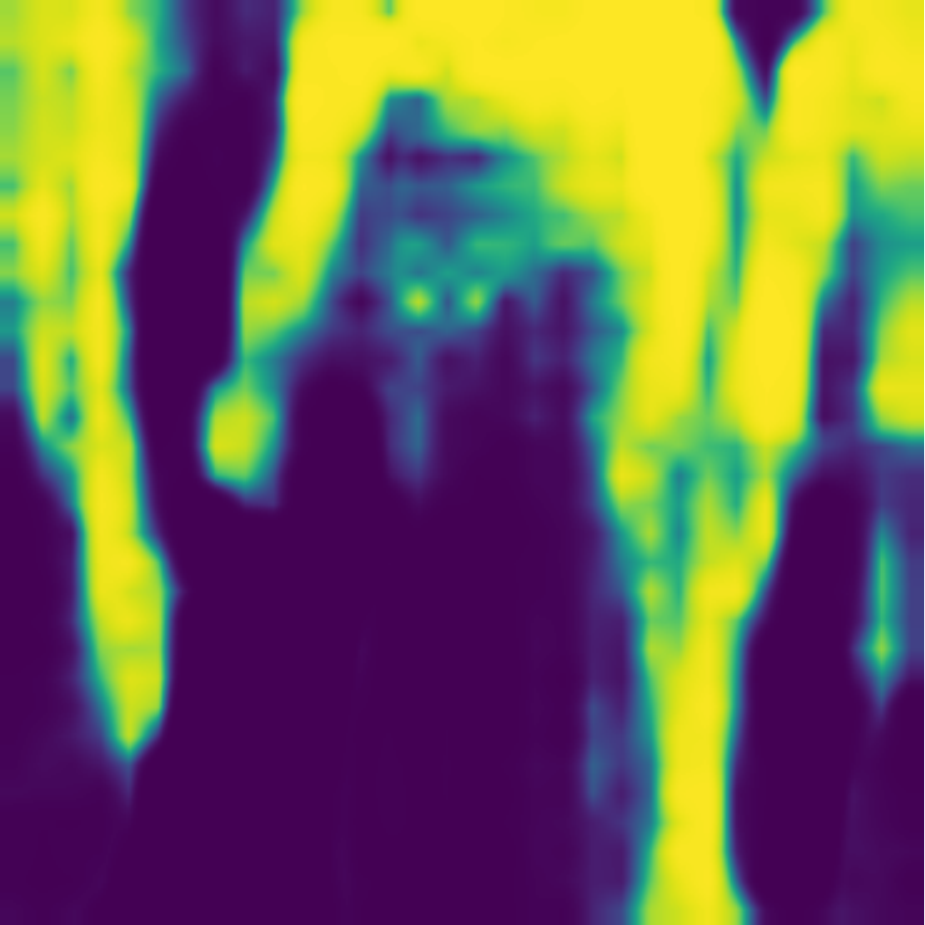}
    }
    \subfloat[\label{fig:inferred:mib_pred} MiB - Pred.]{
    \includegraphics[width=0.19\linewidth]{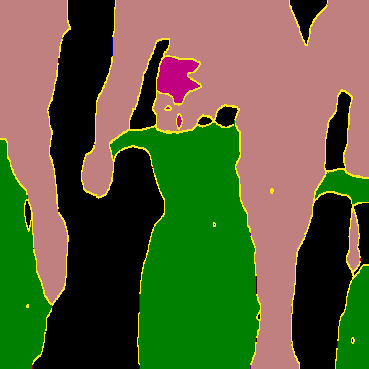}
    }
    \caption{Qualitative example comparing our proposed method to explicit background modeling, MiB \cite{Cermelli_2020_CVPR}.
    }
    \label{fig:inference:prob}
\end{figure*}

\section{Method}
\label{sec:method}
The Semantic Segmentation task consists of predicting a pixel-wise semantic classification from a set of known classes, $\mathcal{C}$, i.e. given an RGB image $\mathcal{X} \in \mathrm{R}^{H \times W \times 3}$ the model should correctly predict the label image $\mathcal{Y} \in \mathcal{C}^{H\times W}$.

Class-Incremental Semantic Segmentation (CSS) aims at incrementally expanding the known set of classes at step $t$, $\mathcal{C}^t$, with new classes $\mathcal{C}^{t+1}$ so that the model learns the full class set $\mathcal{C}^{0:t} := \bigcup_{k=0}^t \mathcal{C}^k$.
However, while \textbf{standard} Semantic Segmentation datasets contain all known classes, fully labeled, CSS only provides labels for the current class set, $\mathcal{C}^t$, and all previous and future class sets are labeled as background.
This introduces a drift in the appearance of the background known as \textit{background shift}.
While recent methods have addressed the background shift \cite{Cermelli_2020_CVPR,douillard2021plop}, their choice to model the background as an explicit class means that future classes are trained to output a high activation for the background class despite it being an incorrect prediction of the completely trained model.

Our proposed method uses implicit modeling of the background class, avoiding that the background is classified as foreground, while not constraining the future class labels.%

\subsection{Implicit background modeling}

In this paper, we formulate the background class as the model dependent epistemic uncertainty estimate based on EDL, avoiding large updates of the background class and the assumption that no future classes are part of the background. %

Formally, we learn a model, $\mathcal{M}$, which predicts the per-pixel class-evidence, $\mathcal{E}$, from an image, $\mathcal{X}$, i.e. $\mathcal{M}: \mathcal{X}\rightarrow\mathcal{E}$. The class evidence, $e$, is used to calculate the concentration parameters, $\alpha$, of the underlying Dirichlet distribution, from which both the uncertainty and the predictive probability are calculated. We directly interpret the epistemic uncertainty as the background probability, i.e. $p_\text{bg} = u$.

\paragraph{Semantic Segmentation using EDL:}
Given $\mathcal{X}$, we predict $e_i$ for each class at each pixel. To interpret these scores as the concentration parameters of a Dirichlet distribution, they must be strictly larger than one.
While the original formulation uses a \text{ReLU} function \cite{sensoy2018evidential}, we found that an exponential term \cite{Evidential_actions_2021_ICPR} performs better. Based on the results in our ablation study (see \cref{tab:abl:exp_sig}) we further found that combining an exponential function with a sigmoid, $\sigma$, to suppress the response from negative activation, performed best. Based on these findings, we calculate the concentration parameters of the Dirichlet distribution, $\alpha$, according to 
\begin{equation}
    \alpha_i = \exp(e_i) \sigma(e_i) + 1 .
\end{equation}

From the Dirichlet parameters, the foreground probabilities and the epistemic uncertainties are calculated according to \ref{eq:belief} and \ref{eq:prob}.
The final predictive probability is factorized into the uncertainty $u$, corresponding to the probability of \textit{not} being foreground (i.e. background probability), and the foreground probabilities, $p_i^\text{fg}$. To calculate the final foreground probabilities, the probability of \textit{not} being background and the individual foreground probabilities are multiplied as follows,

\begin{equation}
    p_i = 
    \begin{cases}
        u, & i=0 (\text{bg}) \\
        (1-u) p_i^\text{fg} & else
    \end{cases} .
\end{equation}

\subsection{Training}
The training is separated into multiple sections for learning new classes, maintaining old class knowledge and balancing the increments. To furhter highlight the modeling part of our method and facilitate comparison with similar works, we keep the training as straightforward as possible.

\paragraph{Learning new classes:} The training objective for learning new classes is the pixel-wise cross-entropy loss over the currently active classes, $\mathcal{C}^t$.

\begin{equation}
    \mathcal{L}_{\text{new}} = - \sum_{i\in\mathcal{C}^t} y_i \log p_i,
    \label{eq:CE_new}
\end{equation}

where $y$ is the one-hot encoded ground-truth label for a certain pixel. The final loss is averaged over all pixels.

\paragraph{Output-level Knowledge Distillation:} 
The learned class knowledge is maintained by two output-level knowledge distillation losses \cite{hinton2015distilling} between the current probability, $p^{(t)}$, and the output from the teacher model trained on the previous increment, $p^{(t-1)}$. The first loss is the foreground multinomial cross-entropy loss that maintains the intraclass variations (\ref{eq:KD_fg}).
The second loss is a binary cross-entropy loss, which maintains the uncertainty with respect to the previous classes (\ref{eq:KD_u}). These two knowledge-distillation terms works as regularization terms to maintain the hidden knowledge of the previous model. 

\begin{equation}
    \mathcal{L}_{\text{KD-fg}} = - \sum_{i\in\mathcal{C}^{1:t-1}} p^{\text{fg,}(t-1)}_i \log p^{\text{fg,}(t)}_i
    \label{eq:KD_fg}
\end{equation}

\begin{equation}
    \mathcal{L}_{\text{KD-u}} = - u^{(t-1)} \log u^{(t)} + (1-u^{(t-1)}) \log (1 - u^{(t)})
    \label{eq:KD_u}
\end{equation}

\paragraph{Increment balancing:}
While the probabilities received by using softmax are independent of the number of classes, the formulation of EDL introduces a dependency between the number of classes and the final probabilities.
To maintain the predictive foreground probability when adding new classes, the activation of each increment needs to be scaled dependently on the number of classes in the step compared to the total number of classes. 
Since the final number of classes is not necessarily known during the beginning of training, we formulate this compensation term for inference only.

The scaling is applied to the maximum activation in each increment that is larger than zero to not scale negative activations.
By requiring the intraclass set probability and the probability after scaling to be the same (the derivation can be found in the supplementary material), we get the following scale factor,
\begin{equation}
    o = \frac{(2 K^t - 1)(K^T)^2}{(K^t)^2(2 K^T - 1)} .
    \label{eq:step_balancing}
\end{equation}

We apply this increment balancing term to all ADE20k \cite{zhou2019semantic} results because of the large number of classes and the difference between the number of base classes and incrementally added new classes.

\paragraph{Foreground-background balancing:}
Compared to Pascal VOC, the classes of ADE20k varies much more in size (average number of pixels per image). For the CISS setting this leads to large variations of the foreground-background ratio of the different increments, especially for certain class-orders.
To adress this, we apply a class-balancing term between foreground and background regions by weighting the loss for the corresponding set of pixels according to the inverse ratio of pixels in the current batch. For training stability, we empirically found that clamping this weight factor to a maximum weight of 10 worked well, i.e. the weight is calculated as

\begin{equation}
    w_i = \min \left( 10, \begin{cases}
     \frac{N_\text{pixels}^\text{tot}}{N_\text{pixels}^\text{bg}} & i=0 \\
     \frac{N_\text{pixels}^\text{tot}}{N_\text{pixels}^\text{fg}} & \text{else}
    \end{cases} \right).
\end{equation}

\section{Experiments}
\label{sec:experiments}
We compare our method with recent state-of-the-art CSS methods on two datasets, the Pascal VOC 2012 \cite{pascal-voc-2012} and the ADE20k \cite{zhou2019semantic} datasets. Multiple variants of the incremental learning task have been proposed in terms of the number of incremental steps, the number of classes in each step, and which classes that might be unlabeled in the background at each step. 
The \textit{overlapped} scenario is the most realistic one, allowing both future and past classes to be labeled as background. %
This is what would be the usual case if the data collection is done incrementally since the data will contain classes that might be later annotated. 
The \textit{disjoint} scenario is arguable less common practical use cases but is more challenging in terms of maintaining old class knowledge.

From a practical perspective, the overlapped scenario contains more data samples per increment since all images showing any of the current labels are used. In the disjoint scenario, only images that do not contain any of the future classes are used, which means that none of the samples will be reused, see \cref{tab:settings} for a comparison.

It is worth noting that ADE20k is only evaluated on the pseudo-disjoint setting proposed by \cite{Cermelli_2020_CVPR}. This setting is the same as a disjoint-setting but with the additional requirement that a minimum number of images of each class is included in each increment.

\begin{table}[h]
    \centering
    \begin{tabular}{c|c|c|c}
                & Bg can contain & Bg can contain & Reused images w/  \\
        Setting & prev. classes & future classes & different label \\
        \hline
        Overlapped & Yes & Yes & Yes \\
        Disjoint & Yes & No & No \\
    \end{tabular}
    \caption{Description of the two setting used on Pascal VOC}
    \label{tab:settings}
\end{table}

\paragraph{Formal problem definition}:
Let the total number of increments be $T$ and a specific increment, $t$, while the current set of classes is notated as $\mathcal{C}^t$ and the current training set as $\mathcal{D}^t$. 
In this case, all sets of classes are disjoint, i.e.
\begin{equation}
    \mathcal{C}^i \cup \mathcal{C}^j = \emptyset, i\neq j .
\end{equation}
All labels in the dataset are part of the current class set, $\mathcal{Y}^t \in \mathcal{C}^t$, with previous and future classes labeled as \textit{background}.

In addition to these class-incremental settings, the \textit{joint} setting is the standard semantic segmentation setting in which all classes are learned in a single increment.

\subsection{Evaluation}
Our method is quantitatively evaluated using seven different setups \cite{Cermelli_2020_CVPR} on two datasets, the Pascal VOC 2012 \cite{pascal-voc-2012} and the ADE20k \cite{zhou2019semantic}, listed in \cref{tab:task_increments}. 

\paragraph{Pascal VOC}
is a widely used dataset for semantic segmentation, which contains 20 annotated foreground classes and one background class for unannotated pixels. The foreground objects are all \textit{things}, i.e \textit{person}, \textit{car}, and \textit{airplane}.
We evaluate our method following four of the PASCAL setups in the evaluation protocol \cite{Cermelli_2020_CVPR}. These consist of a shorter incremental setup with a single step, \textit{15-5}, where the first 15 and the last 5 classes are learned in different steps, and the more complicated \textit{15-1} setup where the last 5 classes are learned separately in different steps, resulting in a total of 6 steps. Both of these setups are evaluated in the \textit{overlap} and \textit{disjoint} settings.

\paragraph{ADE20k}
is, compared to the Pascal dataset, a more challenging dataset with 150 classes of both \textit{thing} classes and \textit{stuff} classes (i.e. \textit{grass}, \textit{sky}) but without an explicit background class. The challenging nature can be seen from the fact that the current state-of-the-art of semantic segmentation achieves a mean Intersection-over-Union (mIoU) score of $53.5$ on the validation dataset \cite{liu2021swin} while it is $90.5$ on Pascal \cite{zoph2020rethinking}.
While the Pascal dataset is reasonably class-balanced, ADE20k is a class-imbalanced dataset, since it contains both large \textit{stuff} classes (i.e. \textit{building}) and smaller \textit{thing} classes (i.e. \textit{fan}). This class imbalance has a major impact on how well the different increments can be learned. Because of this, the ADE20k evaluation protocol \cite{Cermelli_2020_CVPR} evaluates each scenario on two different class orders, the original one in decreasing frequency order and a random order.

We follow the evaluation protocol proposed by \cite{Cermelli_2020_CVPR} and evaluate on three distinct scenarios, \textit{100-50} (2 steps), \textit{100-10} (6 steps), and \textit{50-50} (3 steps). Similar to the naming of the Pascal tasks, the first number is the number of classes in the base step and the second is the number of classes in each increment after that until all 150 classes have been learned.

While PLOP \cite{douillard2021plop} evaluates on ADE20k, they do not evaluate their method on multiple splits and do not use the same disjoint setting as MiB \cite{Cermelli_2020_CVPR}. 
Therefore, the numbers in \cite{douillard2021plop} cannot be compared to the results in \cite{Cermelli_2020_CVPR,Michieli_2021_CVPR} and here.

\begin{table}[h]
    \centering
    \begin{tabular}{ |p{4cm}|p{1.0cm}|p{1cm}|p{1cm}|p{1cm}|p{1cm}|} %
        \hline
        Dataset & \multicolumn{2}{c|}{Pascal} & \multicolumn{3}{c|}{ADE20k}\\
        \hline
        Task & 15-5 & 15-1 & 100-50 & 50-50 & 100-10 \\
        \hline
        Number of  Increments & 2 & 6 & 2 & 3 & 6 \\
        \hline
        New classes per increment & 5 & 1 & 50 & 50 & 10 \\
        \hline
    \end{tabular}
    \caption{Table illustrating some of the properties of the different tasks.}
    \label{tab:task_increments}
\end{table}

\paragraph{Ablation study}
We perform an ablation study on the choice of the rectification function on the Pascal 15-5 overlap scenario, the choice of EDL for implicit background modeling, and the importance of class balancing on ADE20k 100-10 in \cref{sec:ablation}. 

\subsection{Metrics}
The evaluation metric for ADE20k and Pascal is the class-wise mean Intersection-over-Union (mIoU), which is also the metric that previous work reports. However, the standard mIoU metric does not fully illustrate the class-incremental performance of the model. To clearer show how the performance differs between different increments, we report three different mIoU metrics as previously proposed \cite{Cermelli_2020_CVPR}:

\begin{itemize}
    \item Base: mIoU of the base classes trained in the first increment (Pascal: 0-15)
    \item New: mIoU of the incrementally added classes (Pascal: 16-20). 
    \item All: mIoU of all classes including background (Pascal: 0-20)
\end{itemize}

While the mIoU-metric for \textit{all} is of interest for the final system performance, it can often be skewered and obscuring if the model sacrifice the performance of the new classes to maintain the performance of the base classes. %
In addition to the three metrics above, we also propose to use the incremental-wise averaged mIoU, incremental-mIoU, which puts equal importance on all incremental steps independently of the number of classes in each step. This metric differs from the \textit{averaged} metric proposed in \cite{douillard2021plop}, which calculates the averaged mIoU of the performance after each incremental step. 

\begin{equation}
    \text{Inc-mIoU} = \frac{1}{T} \sum_{t=1}^T \frac{1}{\mathcal{C}^t} \sum_{k \in \mathcal{C}^t} IoU_k = \frac{1}{T} \sum_{t=1}^T \text{mIoU}(\mathcal{C}^t).
\end{equation}

\subsection{Implementation details}
We use the same network architecture as previous works (\cite{Cermelli_2020_CVPR,Michieli_2021_CVPR}), which is the DeepLab v3 \cite{chen2017rethinking} with a ResNet101 backbone. 
To facilitate the comparison, we are using the same pretrained weights, the same learning rates, and the same optimizer and learning rate scheduler as \cite{Cermelli_2020_CVPR}, that is, a polynomial learning rate scheduler with a decay rate of $0.9$, a learning rate of $\lambda_0 = 0.01$, and a learning rate of the following steps of $\lambda_{t>0} = 0.001$. The optimizer is the Stochastic Gradient Descent (SGD) with a Nesterov momentum of $0.9$.

The output stride during testing is kept the same as during training, e.g., 16, which means that the final class scores are upsampled from $32\times32$ to $512\times 512$ using bilinear interpolation. We used a batch size of 20 and 8, respectively, and used 30 epochs for Pascal and 60 epochs for training each increment on ADE20k \cite{zhou2019semantic}. No early stopping was employed.
As in previous works we use random crop and horizontal flips as data augmentation, cropping the images randomly to $512\times512$ during training and during validation and testing, using a center-crop of the same size. The relative weighting between the cross-entropy loss and the knowledge distillation loss is the same as in \cite{Cermelli_2020_CVPR,Michieli_2021_CVPR}, $\lambda_{CE} = 1$ and $\lambda_{KD} = 10$.

\section{Results}

We compare our method with the state-of-the-art on Pascal VOC 2012 and ADE20k according to the experimental setup previously described.

\paragraph{Pascal VOC:}
\Cref{tab:voc_results} shows quantitative results on the 15-5 and the 15-1 setups. All of our results are reported as the mean and variance over three distinct random seeds ($42$, $1337$, and $2001$) while the compared methods have used a single seed for evaluation.
In \Cref{tab:voc_results} we can see that our methods outperform all previous methods on Pascal \textit{15-5}.
The results from the more challenging \textit{15-1} scenario (see \cref{tab:voc_results}) also show that our method outperforms most methods except for RECALL \cite{Maracani_2021_ICCV} which uses additional unlabeled data during training.

\begin{figure*}[b!]
    \centering
    \subfloat[\label{fig:scooter:rgb} RGB image]{
    \includegraphics[width=0.24\linewidth]{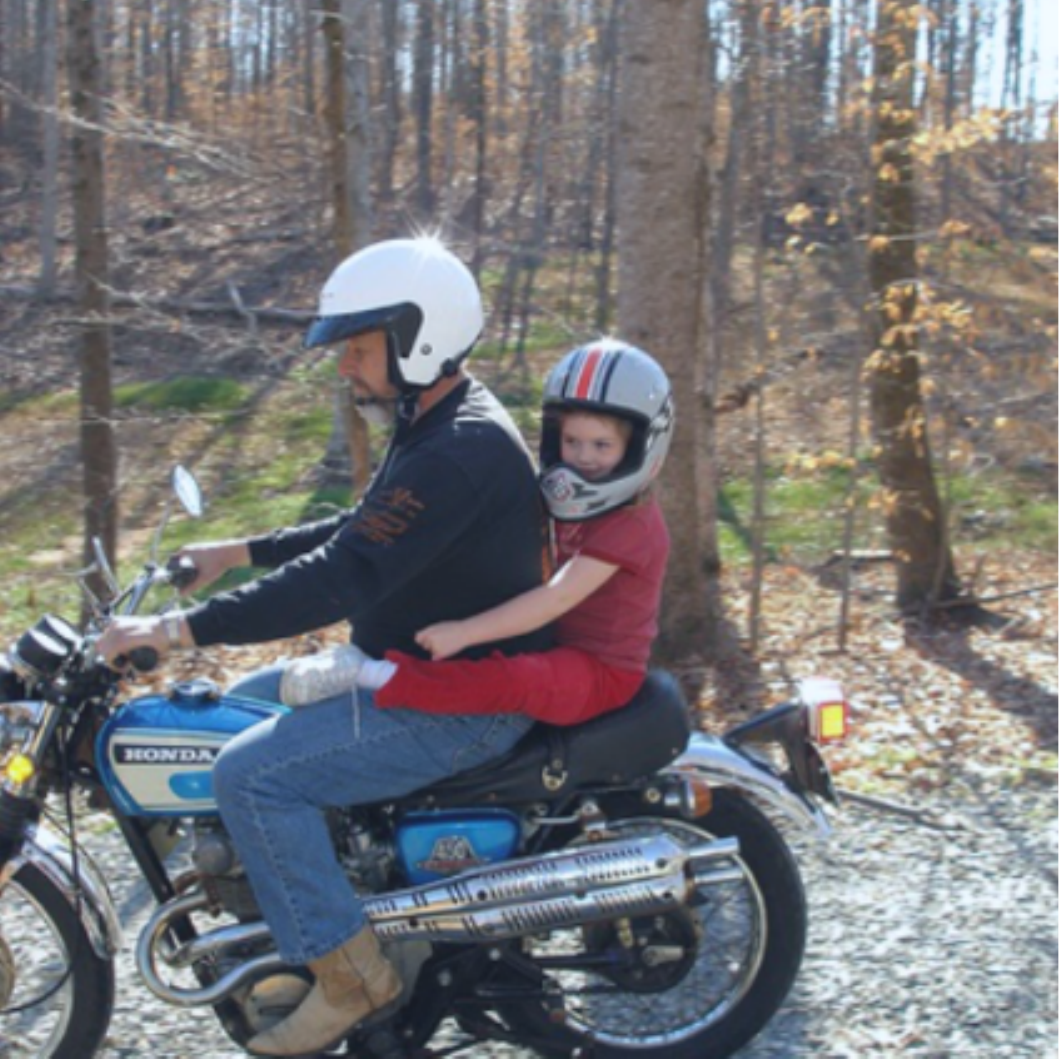}
    }
    \subfloat[\label{fig:scooter:gt} GT labels]{
    \includegraphics[width=0.24\linewidth]{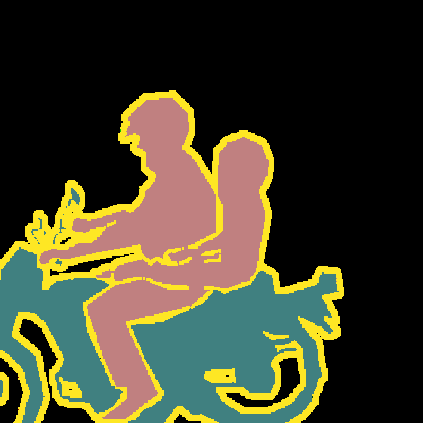}
    }
    \subfloat[\label{fig:scooter:EDL} Our]{
    \includegraphics[width=0.24\linewidth]{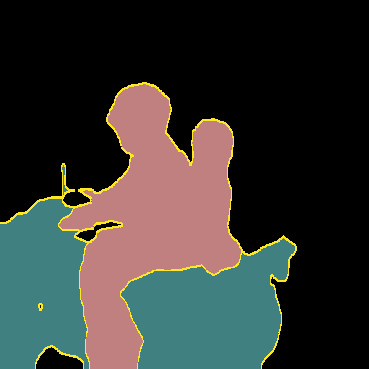}
    }
    \subfloat[\label{fig:scooter:MiB} MiB]{
    \includegraphics[width=0.24\linewidth]{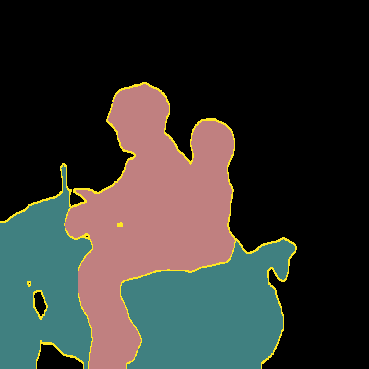}
    }\\
    \subfloat[\label{fig:horse:rgb} RGB image]{
    \includegraphics[width=0.24\linewidth]{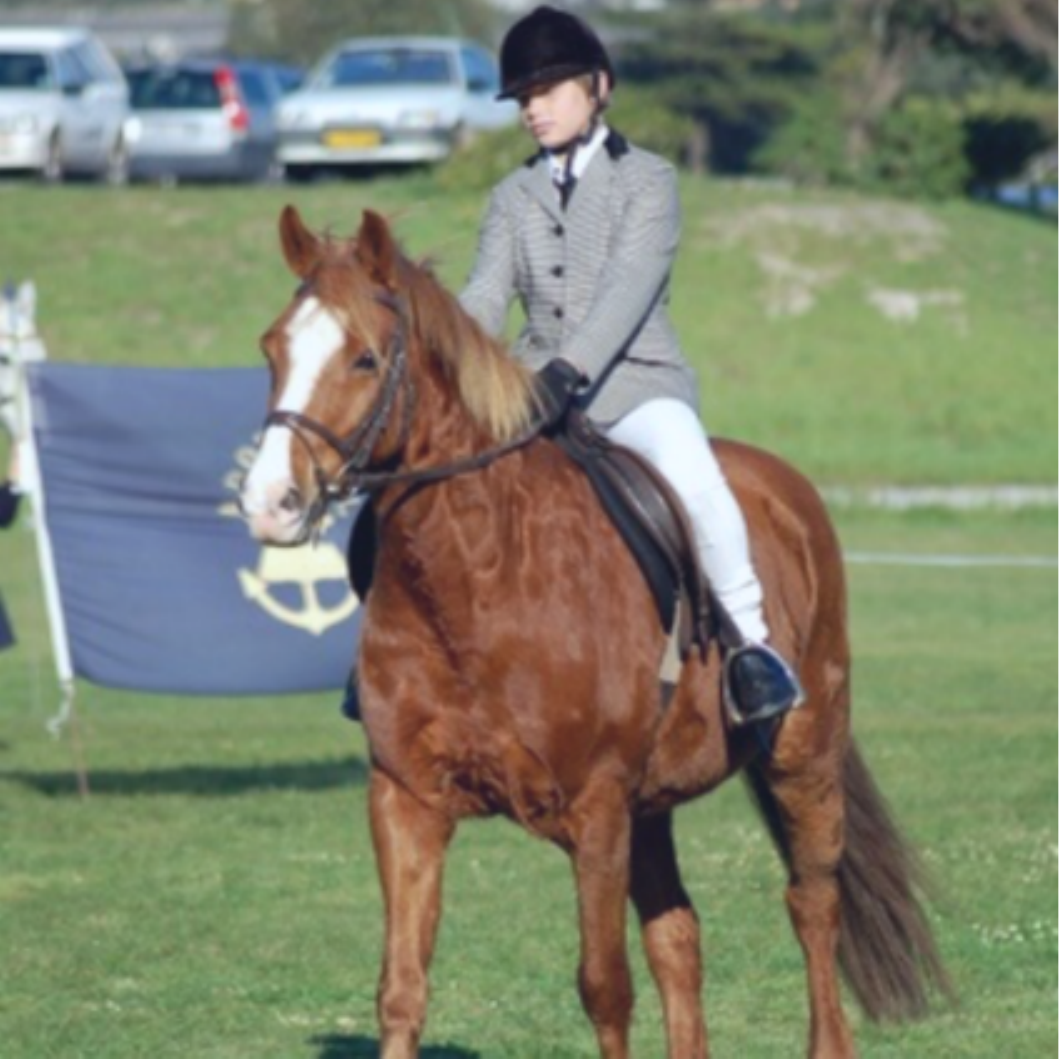}
    }
    \subfloat[\label{fig:horse:gt} GT labels]{
    \includegraphics[width=0.24\linewidth]{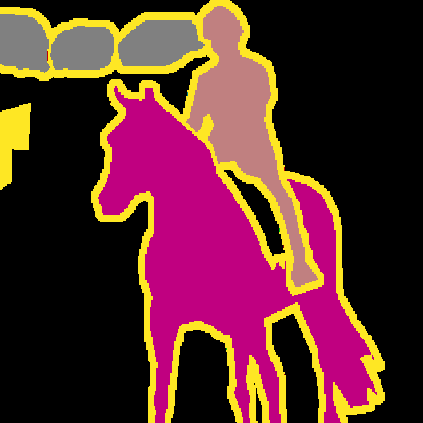}
    }
    \subfloat[\label{fig:horsw:EDL} Our]{
    \includegraphics[width=0.24\linewidth]{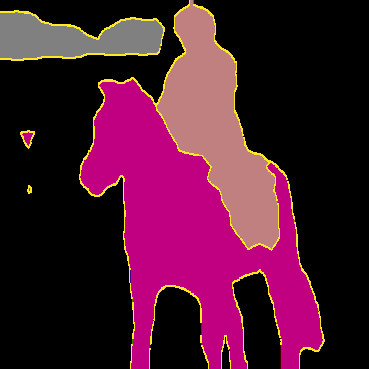}
    }
    \subfloat[\label{fig:horse:MiB} MiB]{
    \includegraphics[width=0.24\linewidth]{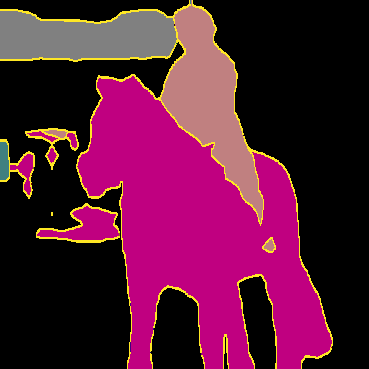}
    }
    \caption{Illustration of quantitative results of our method on Pascal \textit{15-5} compared to MiB \cite{Cermelli_2020_CVPR}}
    \label{fig:interference:pascal:15_5}
\end{figure*}

\begin{figure*}[b!]
    \centering
    \subfloat{
    \includegraphics[width=0.12\linewidth]{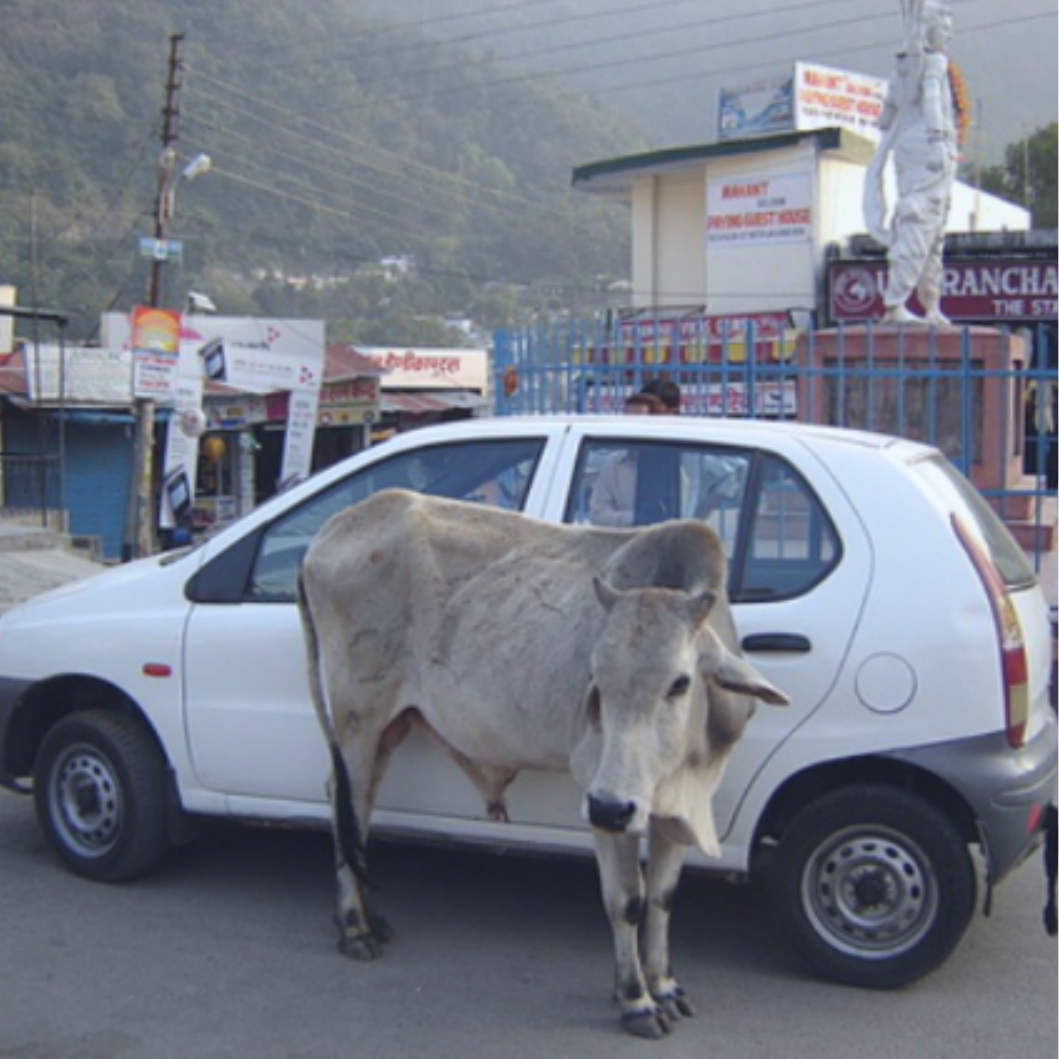}
    }\hspace{-0.3cm}
    \subfloat{
    \includegraphics[width=0.12\linewidth]{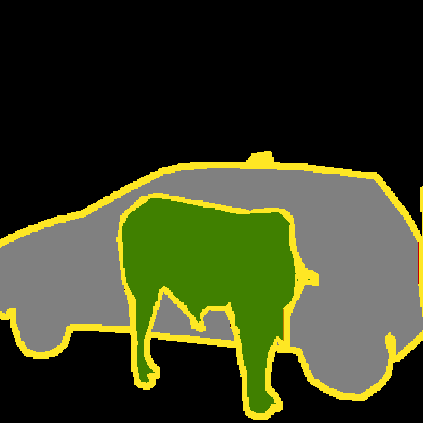}
    }\hspace{-0.3cm}
    \subfloat{
    \includegraphics[width=0.12\linewidth]{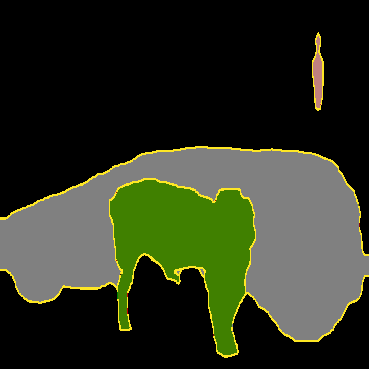}
    }\hspace{-0.3cm}
    \subfloat{
    \includegraphics[width=0.12\linewidth]{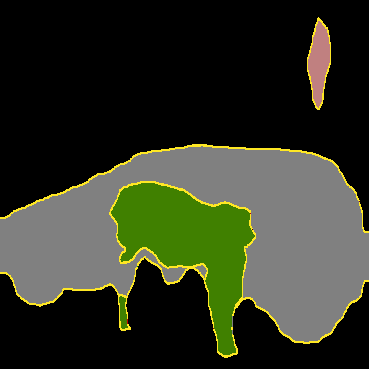}
    }\hspace{-0.3cm}
    \subfloat{
    \includegraphics[width=0.12\linewidth]{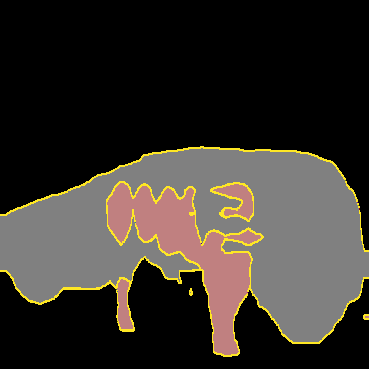}
    }\hspace{-0.3cm}
    \subfloat{
    \includegraphics[width=0.12\linewidth]{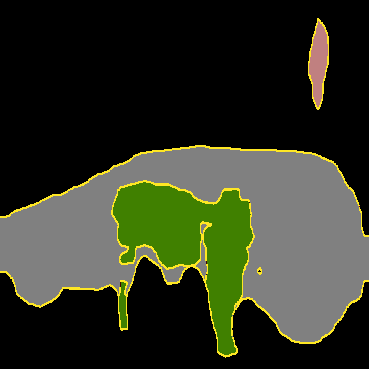}
    }\hspace{-0.3cm}
    \subfloat{
    \includegraphics[width=0.12\linewidth]{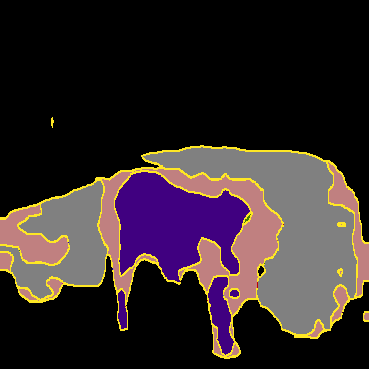}
    }\hspace{-0.3cm}
    \subfloat{
    \includegraphics[width=0.12\linewidth]{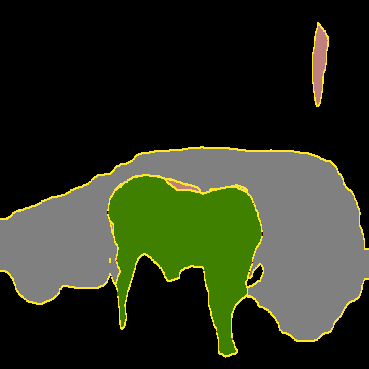}
    }\vspace{-0.3cm}
    \\
    \subfloat{
    \includegraphics[width=0.12\linewidth]{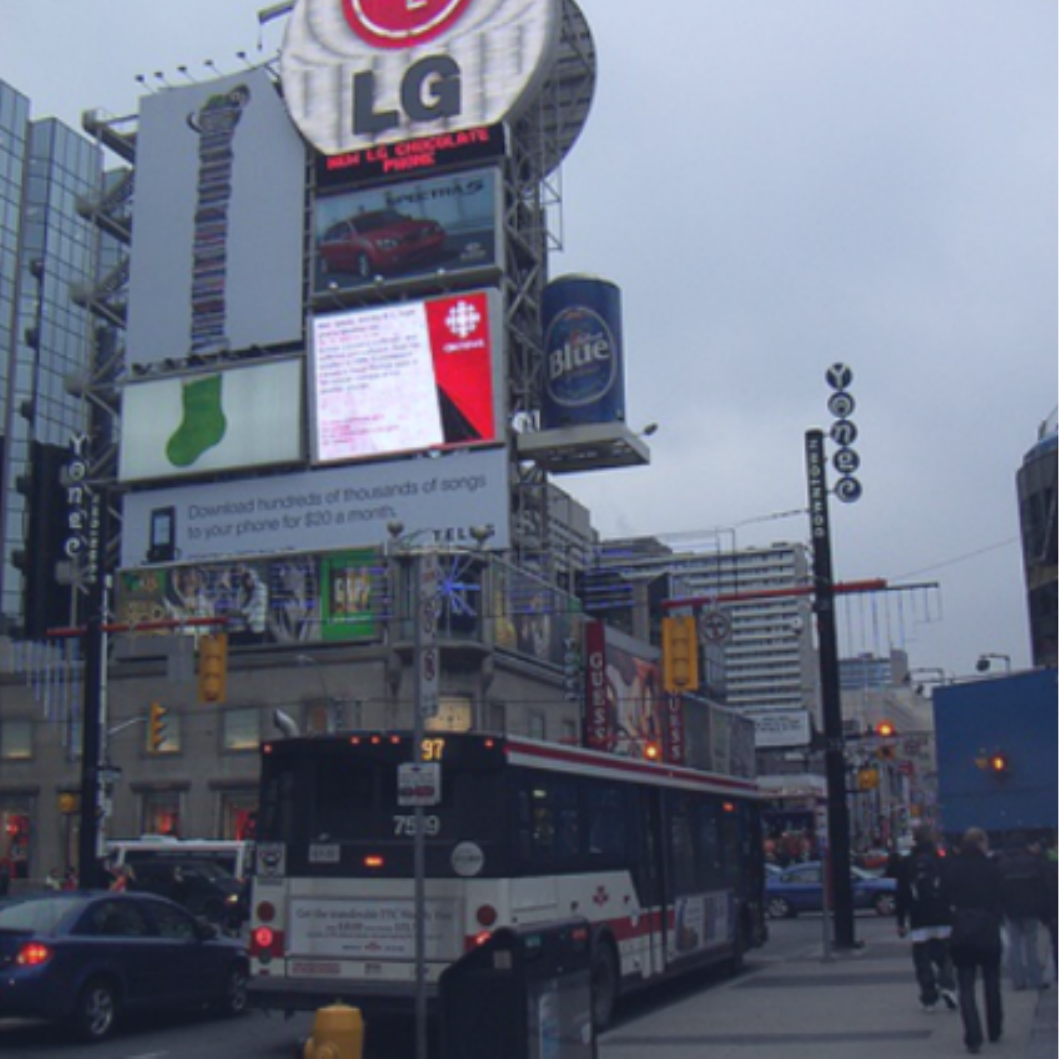}
    }\hspace{-0.3cm}
    \subfloat{
    \includegraphics[width=0.12\linewidth]{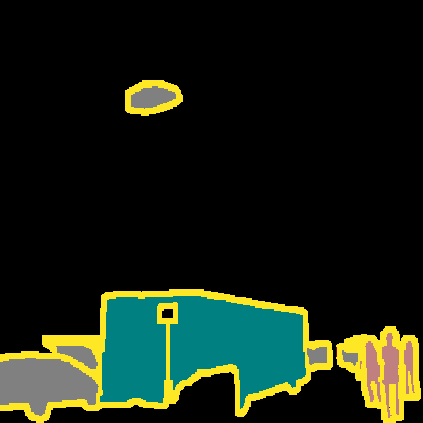}
    }\hspace{-0.3cm}
    \subfloat{
    \includegraphics[width=0.12\linewidth]{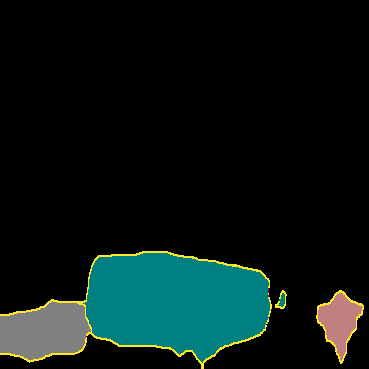}
    }\hspace{-0.3cm}
    \subfloat{
    \includegraphics[width=0.12\linewidth]{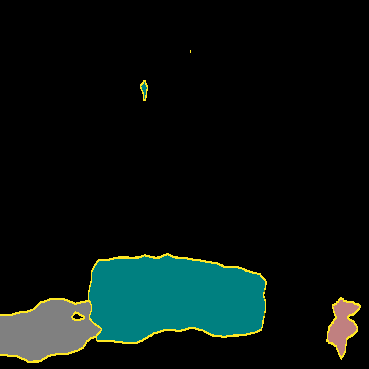}
    }\hspace{-0.3cm}
    \subfloat{
    \includegraphics[width=0.12\linewidth]{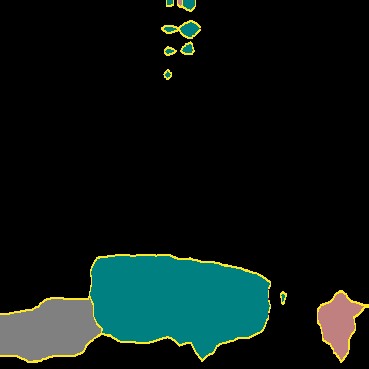}
    }\hspace{-0.3cm}
    \subfloat{
    \includegraphics[width=0.12\linewidth]{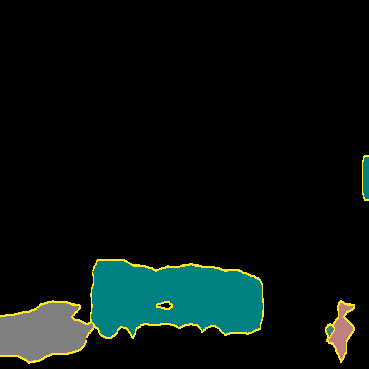}
    }\hspace{-0.3cm}
    \subfloat{
    \includegraphics[width=0.12\linewidth]{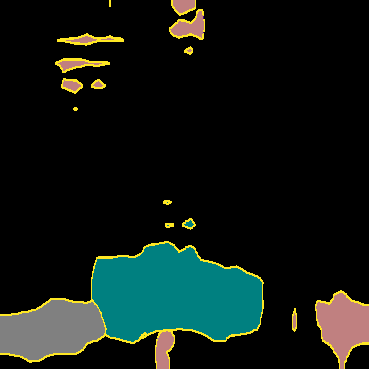}
    }\hspace{-0.3cm}
    \subfloat{
    \includegraphics[width=0.12\linewidth]{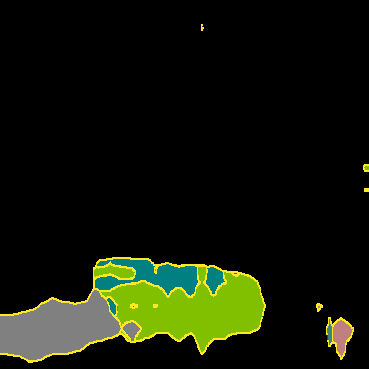}
    }\vspace{-0.3cm}
    \\
    \subfloat[RGB]{
    \includegraphics[width=0.12\linewidth]{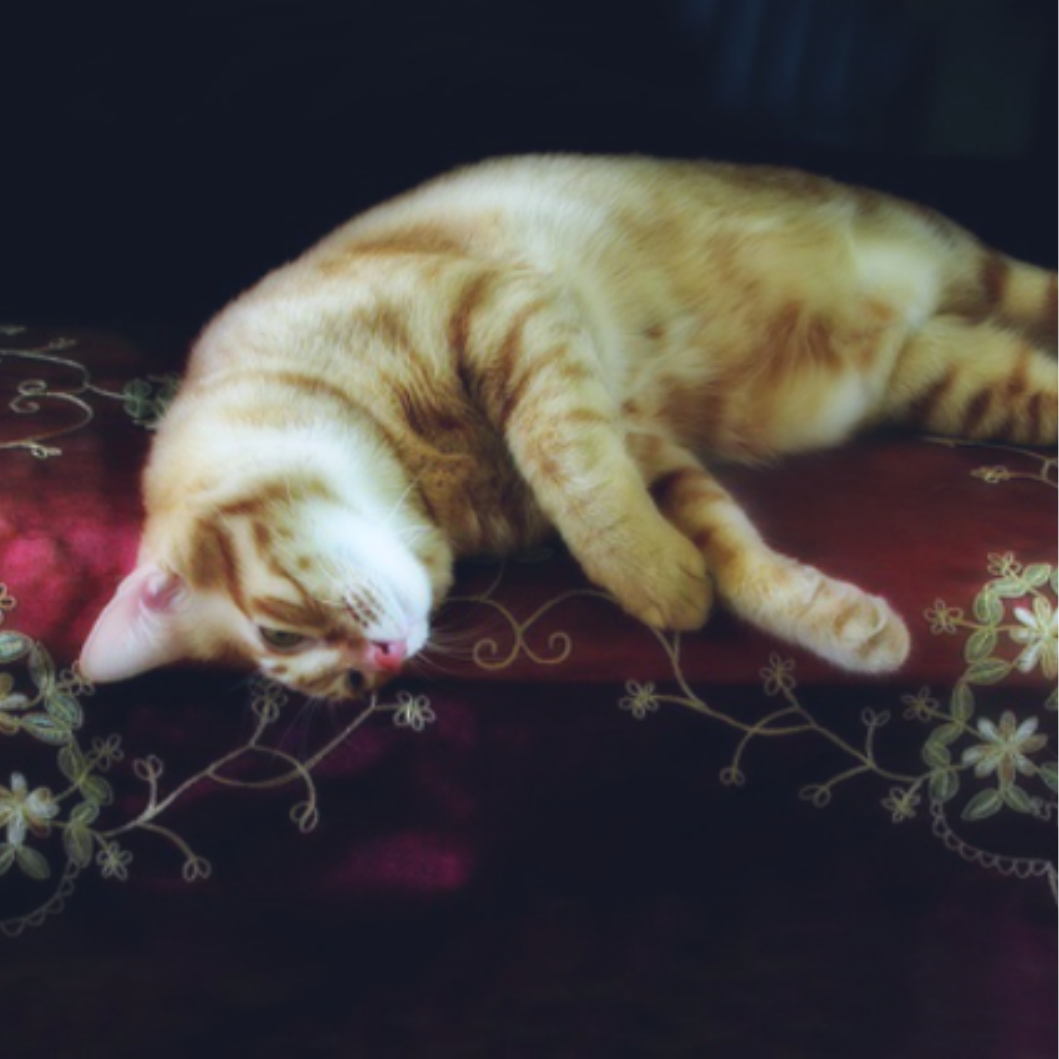}
    }\hspace{-0.3cm}
    \subfloat[GT]{
    \includegraphics[width=0.12\linewidth]{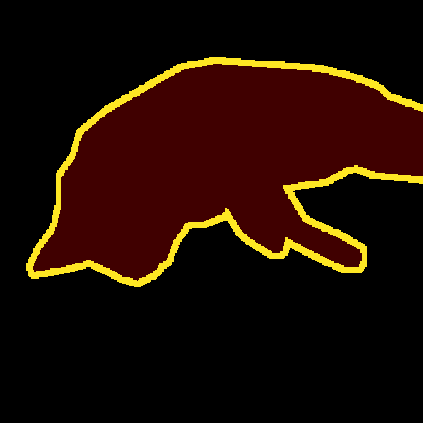}
    }\hspace{-0.3cm}
    \subfloat[0-15]{
    \includegraphics[width=0.12\linewidth]{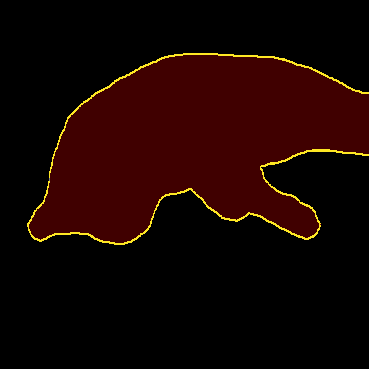}
    }\hspace{-0.3cm}
    \subfloat[\textit{plant}]{
    \includegraphics[width=0.12\linewidth]{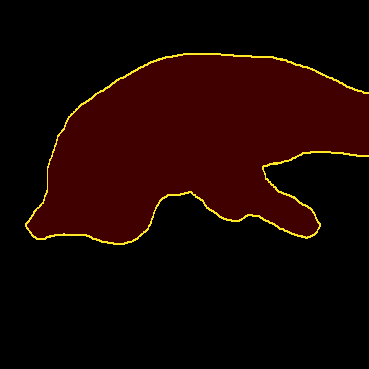}
    }\hspace{-0.3cm}
    \subfloat[\textit{sheep}]{
    \includegraphics[width=0.12\linewidth]{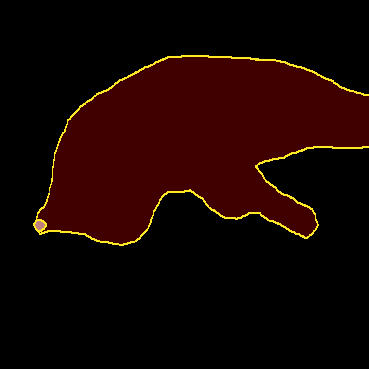}
    }\hspace{-0.3cm}
    \subfloat[\textit{sofa}]{
    \includegraphics[width=0.12\linewidth]{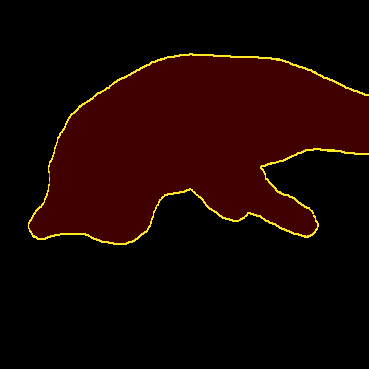}
    }\hspace{-0.3cm}
    \subfloat[\textit{train}]{
    \includegraphics[width=0.12\linewidth]{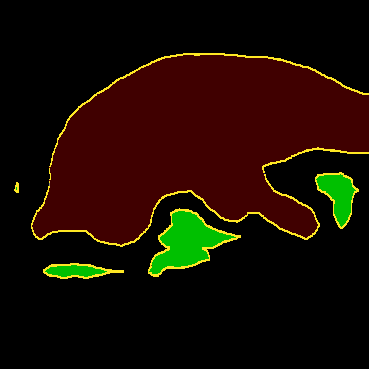}
    }\hspace{-0.3cm}
    \subfloat[\textit{tv}]{
    \includegraphics[width=0.12\linewidth]{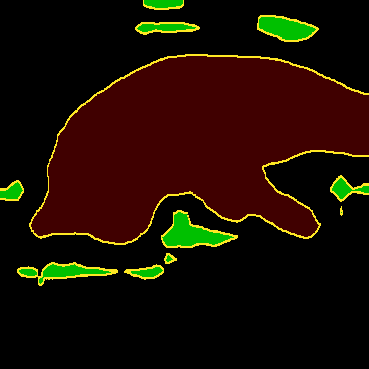}
    }
    \caption{
    Illustration of how the class predictions of previous classes change while learning new classes. The figures show the predictions after each step on Pascal \textit{15-1}. Note the class confusion that occurs right after learning the new class sheep in the top row, where the cow is miss-classified as such. Similar effects occur in the middle row where the bus and train classes are similar.
    Overall, we can see that even in the challenging scenario of Pascal \textit{15-1}, our method maintains high accuracy for the learned classes.
    }
    \label{fig:interference:pascal:15-1}
\end{figure*}

\begin{table*}[h]
    \centering
    \resizebox{\linewidth}{!}{%
    \begin{tabular}{ p{3cm}||p{0.80cm} p{0.80cm} p{0.80cm} p{0.80cm} | p{0.80cm} p{0.80cm} p{0.80cm} p{0.80cm} || p{0.80cm} p{0.80cm} p{0.80cm} p{0.8cm} | p{0.8cm} p{0.8cm} p{0.8cm} p{0.8cm} }
        & \multicolumn{8}{c}{15-5} & \multicolumn{8}{c}{15-1}\\
        & \multicolumn{4}{c|}{disjoint} & \multicolumn{4}{c||}{overlapped} & \multicolumn{4}{c|}{disjoint} & \multicolumn{4}{c}{overlapped} \\
        & 0-15 & 16-20 & All & Inc. & 0-15 & 16-20 & All & Inc. & 0-15 & 16-20 & All & Inc. & 0-15 & 16-20 & All & Inc.\\
        \hline
        MiB \cite{Cermelli_2020_CVPR} & 
        71.4 & 43.3 & 64.7 & 57.4 & %
        75.1 & 49.4 & 69.0 & 62.3 & %
        45.7 & 12.9 & 37.9 & 18.4 & %
        34.8 & 13.5 & 29.7 & 17.1 \\ %
        MiB \cite{Michieli_2021_CVPR} & 
        47.5 & 34.1 & 44.3 & 40.8 & 
        73.1 & 44.5 & 66.3 & 58.8  & 
        39.0 & 15.0 & 33.3 & 19.0 &
        44.5 & 11.7 & 36.7 & 17.2\\
        SDR \cite{Michieli_2021_CVPR} & 
        73.5 & \textbf{47.3} & \textit{67.2} & \textbf{60.4} & 
        75.4 & \textit{52.6} & 69.9 & \textit{64.0} & 
        59.2 & 12.9 & 48.1 & 20.6 & 
        44.7 & \textit{21.8} & 39.2 & 25.6\\
        SDR+MiB \cite{Michieli_2021_CVPR} & 
        \textit{73.6} & 44.1 & \textbf{67.3} & 58.8 &
        \textbf{76.3} & 50.2 & \textit{70.1} & 63.3 & 
        \textit{59.4} & \textit{14.3} & \textit{48.7} & \textit{21.8} & 
        47.3 & 14.7 & 39.5 & 20.1\\
        PLOP \cite{douillard2021plop} &
        - & - & - & - &
        \textit{75.7} & 51.7 & \textit{70.1} & 63.7 &
        - & - & - & - &
        \textit{65.1} & 21.1 & \textit{54.6} & \textit{28.4} \\
        \hline
        RECALL \cite{Maracani_2021_ICCV} \small(GAN)  & 
        67.8 & 49.8 & 63.5 & 58.8 &%
        68.1 & 50.9 & 64.0 & 59.5 &%
        67.5 & 44.9 & 62.1 & 48.7 &%
        69.0 & 49.2 & 64.3 & 52.5 \\ %
        RECALL \cite{Maracani_2021_ICCV} \small(Web)& 
        70.5 & 52.9 & 66.3 & 61.7 &%
        69.1 & 54.3 & 65.6 & 61.7 & %
        67.4 & 47.8 & 62.7 & 51.1 &%
        69.1 & 50.9 & 64.8 & 53.9 \\%$53.7^*$ \\
        \hline
        
        Our &
        \textbf{73.8} \small $\pm 0.1$ &
        \textit{44.8} \small$\pm 0.0.8$ &
        66.9 \small$\pm 0.1$ &
        \textit{59.3} \small$\pm 0.3$ &
        \textbf{76.3}  \small$\pm 0.9$ &
        \textbf{53.6} \small$\pm 7.4$ &
        \textbf{70.9} \small$\pm 1.5$ &
        \textbf{64.9} \small$\pm 2.8$ &
        \textbf{63.8}  \small$\pm 1.6$ &
        \textbf{15.9} \small$\pm 3.8$ &
        \textbf{52.4} \small$\pm 1.0$ &
        \textbf{23.9} \small$\pm 1.6$ &
        \textbf{68.5} \small$\pm 0.3$ &
        \textbf{24.4} \small$\pm 2.6$ &
        \textbf{58.0} \small$\pm 0.03$ &
        \textbf{31.8} \small$\pm 1.6$
        \\
        \hline
        Our (joint) &
        - & - & 78.8 & - &
        - & - & 78.8 & - &
        - & - & 78.8 & - &
        - & - & 78.8 & - \\
    \end{tabular}%
    }
    \caption{Results on the Pascal VOC dataset, best in \textbf{bold} and runner-up in \textit{italic}. 
    The \textit{joint} setting is used as an upper limit of what the model can achieve when learning all classes simultaneously in one step. RECALL \cite{Maracani_2021_ICCV} is part of the table but not part of the ranking since the method uses additional unlabeled data.
    }
    
    \label{tab:voc_results}
\end{table*}

\paragraph{ADE20k:}

The ADE20k results in \Cref{tab:ade_results} show that our method performs on par with most approaches on the \textit{100-50} and \textit{50-50} settings while out-performing all methods on the most challenging incremental setting \textit{100-10}.

\begin{table*}[t]
    \centering
    \resizebox{0.9\textwidth}{!}{%
    \begin{tabular}{ p{2.5cm}||p{0.8cm} p{0.8cm} p{0.8cm} p{0.8cm} || p{0.8cm} p{0.8cm} p{0.8cm} p{0.8cm} || p{0.8cm} p{0.8cm} p{0.8cm} p{0.8cm} }
    & \multicolumn{4}{c||}{100-50} & \multicolumn{4}{c||}{50-50} & \multicolumn{4}{c}{100-10}\\
    & Base & New & All & Inc. & Base & New & All & Inc. & Base & New & All & Inc. \\
    \hline
    MiB \cite{Cermelli_2020_CVPR} & 
    \textbf{37.9} & \textit{27.9} & \textbf{34.6} & \textbf{32.9} & 
    35.5 & 22.9 & 27.1 & 27.1 &
    \textbf{31.8} & \textit{14.1} & \textit{25.9} & \textit{17.0} \\
    MiB \cite{Michieli_2021_CVPR} & 
    \textit{37.6} & 24.7 & 33.3 & 31.2 & 
    39.1 & 22.6 & 28.1 & 28.1 &
    21.0 & 5.3 & 15.8 & 7.9 \\
    SDR \cite{Michieli_2021_CVPR} & 
    37.4 & 24.8 & 33.2 & 31.1 & 
    \textit{40.9} & 23.8 & 29.5 & 29.5 & 
    \textit{28.9} & 7.4 & 21.7 & 11.0 \\
    SDR+MiB \cite{Michieli_2021_CVPR} & 
    37.5 & 25.5 & \textit{33.5} & \textit{31.5} &
    \textbf{42.9} & \textit{25.4} & \textbf{31.3} & \textbf{31.3} & 
    \textit{28.9} & 11.7 & 23.2 & 14.6 \\
    \hline
    Our &
    33.3 & \textbf{29.1} & 31.9 & 31.2 &
    34.4 & \textbf{27.4} & \textit{29.8} & \textit{29.8} &
    28.5 & \textbf{22.6} & \textbf{26.5} & \textbf{23.6} \\
    \hline 
    joint &
    - & - & 41.1 & - &
    - & - & 41.1 & - &
    - & - & 41.1 & - \\
    \end{tabular}%
    }
    \caption{Results on the ADE20k dataset. The methods have been evaluated using two distinct class orders, the original class order, as well as the random order from \cite{Cermelli_2020_CVPR}.  
    The \textit{Base} set correspond to classes 1-100 and 1-50 respectively, the \textit{New} set correspond to the incrementally added classes 51-150 and 101-150 respectively. Class 0, which is seen as the background is absent in the ADE20k since all pixels are annotated.}
    \label{tab:ade_results}
\end{table*}

\paragraph{Ablation study}
\label{sec:ablation}
We evaluate three important components of our method. 

First, we evaluate the choice of rectification function for the evidence score.
Our choice (\cref{tab:abl:exp_sig}) clearly outperforms ReLU and is superior to Exp.

Secondly, we compare our implicit background modeling approach compared to simply replacing the weight layer for the background class in MiB with a single bias value (see \cref{tab:abl:Bg_model}). We also evaluate how freezing the bias term together with the previous weights after learning the first increment affects the performance. The results clearly show that even a simple implicit background model can be beneficial, however, our proposed EDL-based modeling is clearly superior.

Lastly, we evaluate the effect of the compensation term on ADE20k (see \cref{tab:abl:comp}) and show that it is highly beneficial, especially for the improving the Incremental mIoU.

\begin{table}[htpb]
    \centering
    \resizebox{\linewidth}{!}{
    \subcaptionbox{
    \label{tab:abl:exp_sig}
    }[0.18\textwidth]{
    \flushleft
    \begin{tabular}[t]{l|c}
        Func. & All \\
        \hline
        ReLU  & 53.9 \\
        Exp  & 70.6 \\
        Ours & \textbf{70.9} 
    \end{tabular}
    }%
    \subcaptionbox{
    \label{tab:abl:Bg_model}
    }[0.42\textwidth]{
    \begin{tabular}{ll|c}
        \multicolumn{2}{l}{Method} & mIoU All \\
        \hline
        MiB & & 68.8 \\
        + & Bias only & 69.5 \\
        + & Frozen clas. & 69.1 \\
        \hline
        EDL & Impl. bg & \textbf{70.9} 
    \end{tabular}
    }
    \subcaptionbox{
    \label{tab:abl:comp}
    }[0.25\textwidth]{
    \begin{tabular}[t]{l|c|c}
        Balancing & All & Inc. \\
        \hline
        w/o & 25.8 & 15.8 \\
        with & \textbf{26.5} & \textbf{23.6}
    \end{tabular}
    }
    }
    \caption{Ablation study: a) Influence of choice of rectification function on Pascal 15-5 Overlap. b) Implicit background approach c) Increment balancing on ADE20k 100-10}
    \label{tab:ablation}
\end{table}

\section{Discussion}
\paragraph{Quantitative results:}
From the results in \Cref{tab:voc_results} and \Cref{tab:ade_results} we see that our method based on implicit modeling of the background performs better than most methods using explicit background modeling and without additional data.%
The results for ADE20k show that we perform on par with the state-of-the-art on the two simpler scenarios and clearly surpass the results by SDR \cite{Michieli_2021_CVPR} and MiB \cite{Cermelli_2020_CVPR} when using multiple increments.

\paragraph{Qualitative results:}
To illustrate the qualitative results of our method we pick image samples from the test set and visualize them. \Cref{fig:inference:prob} illustrates the class probabilities of our implicit background modeling compared to an explicit background modeling approach \cite{Cermelli_2020_CVPR}. In \Cref{fig:interference:pascal:15-1} we illustrate how the prediction on older classes is affected by learning new ones. It highlights the remaining problem that new classes, similar to previous ones, can cause confusion between the two classes. Some examples of this are \textit{bus} and \textit{train}, and \textit{cow} and \textit{sheep}, all of which can be similar in certain situations.
Finally, we show some results on the Pascal \textit{15-5} task, where we compare our model to MiB \cite{Cermelli_2020_CVPR}.
In the supplementary material, more qualitative examples can be found showing a comparison with additional methods, a comparison between \textit{15-1} and \textit{15-5} results on Pascal VOC, and a comparison between results on split A and B on ADE20k.

To summarize, our method shows that an implicit background model is not only feasible but also leads to improved performance on the Class-incremental Semantic Segmentation problem, in particular for many increments.

\section{Conclusion}
In this paper, we introduce a novel method for implicitly modeling the background, which consists of unlabeled classes. While the typical explicit background modeling is efficient for standard semantic segmentation, we argue that the class-incremental setting implies an open-world assumption. This assumption requires an implicit background model so as not to constrain the feature space of the unlabeled classes unnecessarily.
We have demonstrated the strengths of our method compared to the state-of-the-art on two datasets and shown that we outperform the state-of-the-art when learning an increasing number of increments.

\newpage

\clearpage

\newpage
{\small
\bibliographystyle{splncs04}
\bibliography{egbib}
}

\clearpage
\newpage
\appendix

\vspace{-20pt}
\begin{center}
\textbf{\Large Supplementary Material for}
\end{center}
\begin{center}\textbf{\Large Evidential Deep Learning for Class-Incremental Semantic Segmentation}
\end{center}

\section{Derivation of Increment Balancing Term}

Due to the relatively straightforward derivation and limited space, the derivation of the increment balancing term was omitted from the main paper. Instead, it is provided here for completeness.

The formulation of EDL introduces a dependency between the number of classes and the uncertainty, (3, 4) in the main paper. Compensating for the bias that occurs between increments in necessary since the number of classes varies over the increments. This is why a post-processing step to rebalance the logits of the different increments was proposed in (10) in the main paper.
Under the assumption of perfect classification of the network, all evidence is centered on the correct class $j$, the complete class probability is written as,
\begin{equation}
    p_i^{(t)} = p_i^{\text{fg},(t)} (1 - u^{(t)}) =  \frac{e_i + 1}{e_i + K^{(t)}} \left( 1 - \frac{K^{(t)}}{K^{(t)} + e_i} \right).
\end{equation}

Where $K^t$ is the number of classes in increment $t$, i.e. $|\mathcal{C}^t|$.
To maintain the foreground probability constant between the training of as single increment and during the inference oveer all learned classes, the class evidence are scaled differently based on the number of classes in the corresponding increment.
That is, the unscaled single increment probability should be equal to the rescaled full probability when considering all classes, i.e.
\begin{equation}
    \frac{p_i^{(t)}}{p_i^{(1:T)}} = 1 \Longleftrightarrow \frac{\frac{e_i + 1}{e_i + K^(t)} \left( 1 - \frac{K^(t)}{K^(t) + e_i} \right)}{\frac{o e_i + 1}{o e_i + K^{(1:T)}} \left( 1 - \frac{K^{(1:T)}}{K^{(1:T)} + o e_i} \right)} = 1
\end{equation}

Solving the above equation with respect to the scale factor, $o$, we arrive at the proposed compensation term,

\begin{equation}
    o = \frac{(2 K^t - 1)(K^{1:T})^2}{(K^t)^2(2K^{1:T} - 1)}.
    \label{sup:correction}
\end{equation}

The compensation term is well defined since the number of classes is at least one.

\section{Additional Qualitative results}
In this section, additional qualitative results to Figures 2, 3, and 4 in the main paper are presented.%
Following is a short description of each method in the comparison:

\paragraph{Fine-tuning (FT).}
While fine-tuning has been proven insufficient for class-in\-cre\-mental learning \cite{Cermelli_2020_CVPR}, we include some qualitative results of using fine-tuning together with our implicit background formulation. This is accomplished by training our method with $\lambda_{KD}=0$.

\paragraph{MiB \cite{Cermelli_2020_CVPR}}
 is trained in our framework using the code provided by the authors of MiB to run inference on Pascal VOC.

\paragraph{PLOP \cite{douillard2021plop}}
is trained on the \textit{overlap} scenario of Pascal VOC using the code provided by the authors and the default settings for their methods. Code can be found here: \href{https://github.com/arthurdouillard/CVPR2021_PLOP/tree/0fb13774735961a6cb50ccfee6ca99d0d30b27bc}{CVPR 2021 PLOP}

\subsection{Results on PASCAL}
\begin{figure*}
    \centering
    \subfloat[\label{fig:sup:a} RGB]{
    \includegraphics[width=0.15\linewidth]{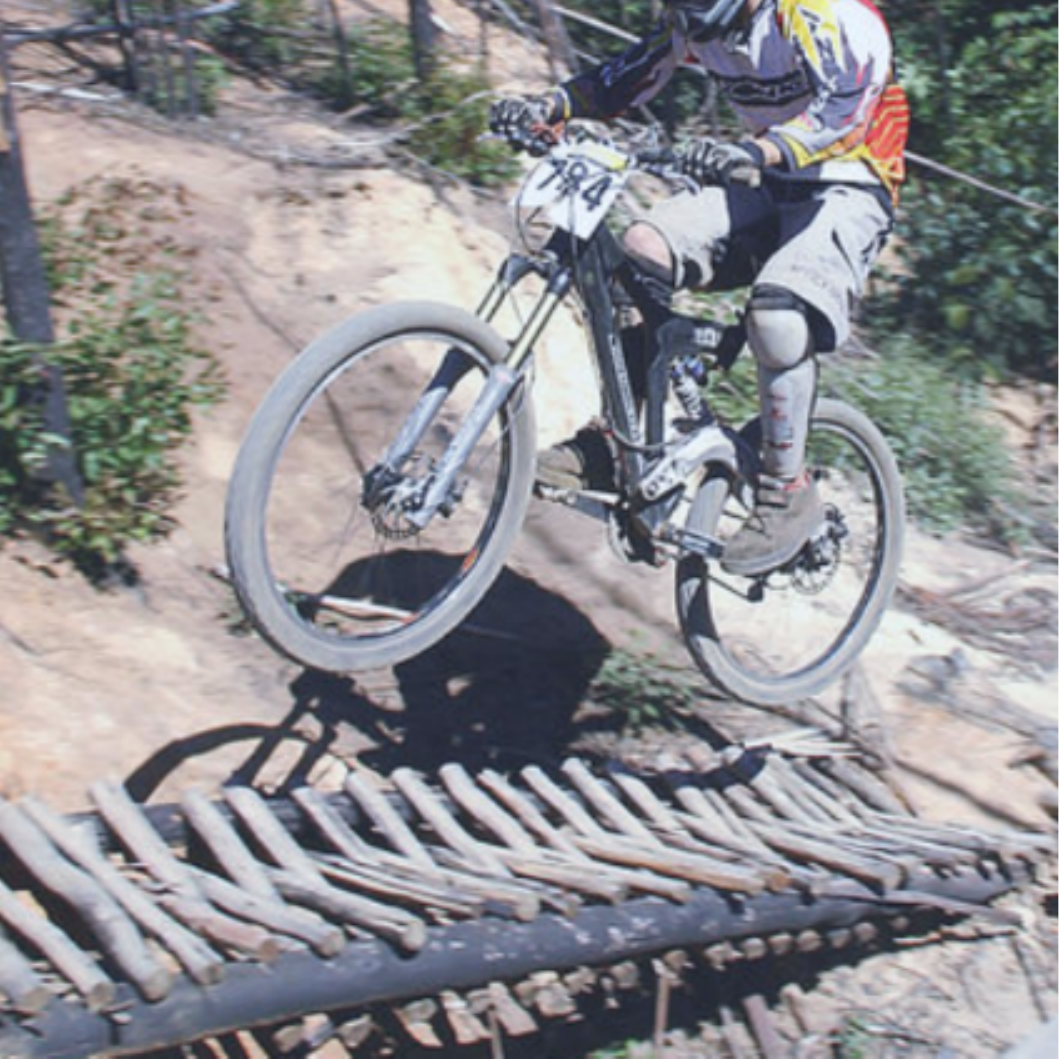}
    }
    \subfloat[\label{fig:sup:b} GT]{
    \includegraphics[width=0.15\linewidth]{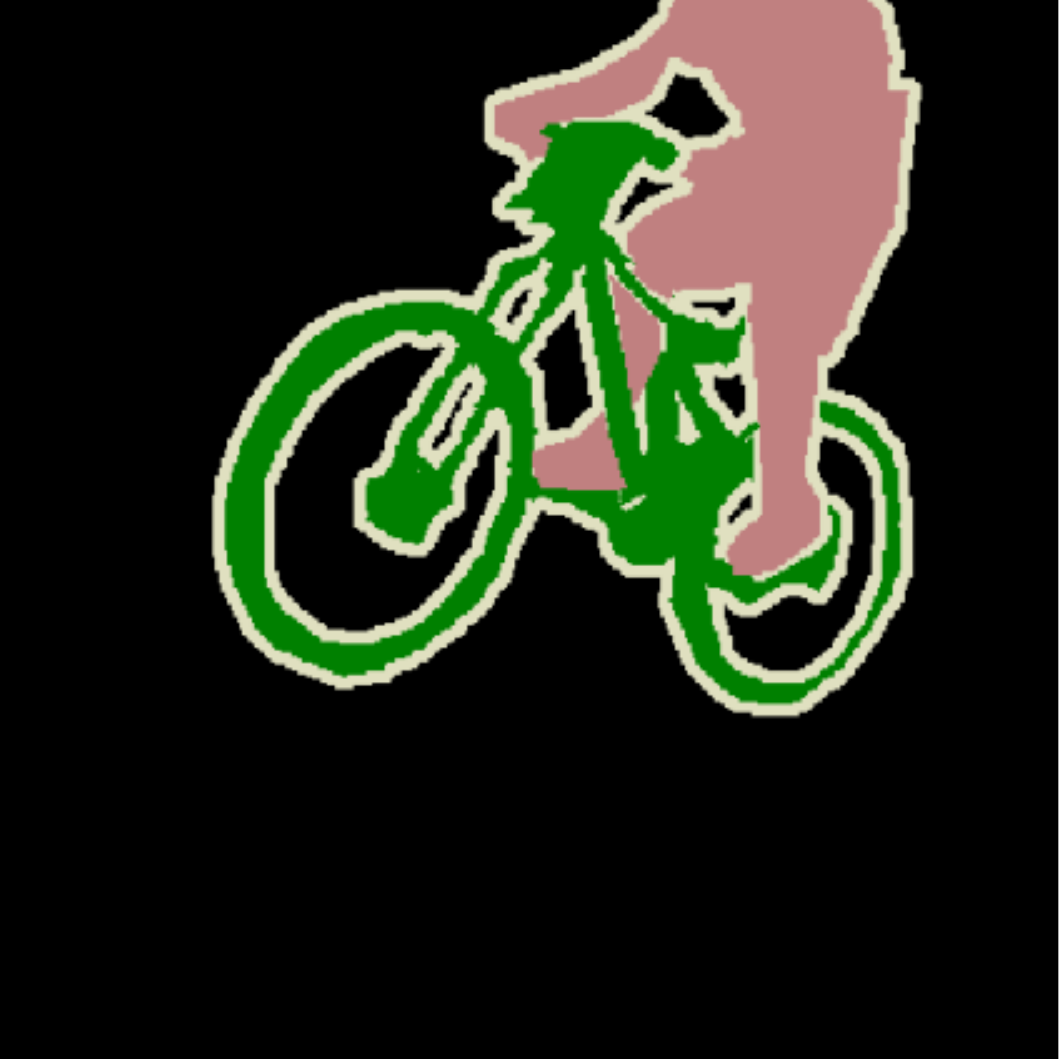}
    }
    \subfloat[\label{fig:sup:c} FT (ours)]{
    \includegraphics[width=0.15\linewidth]{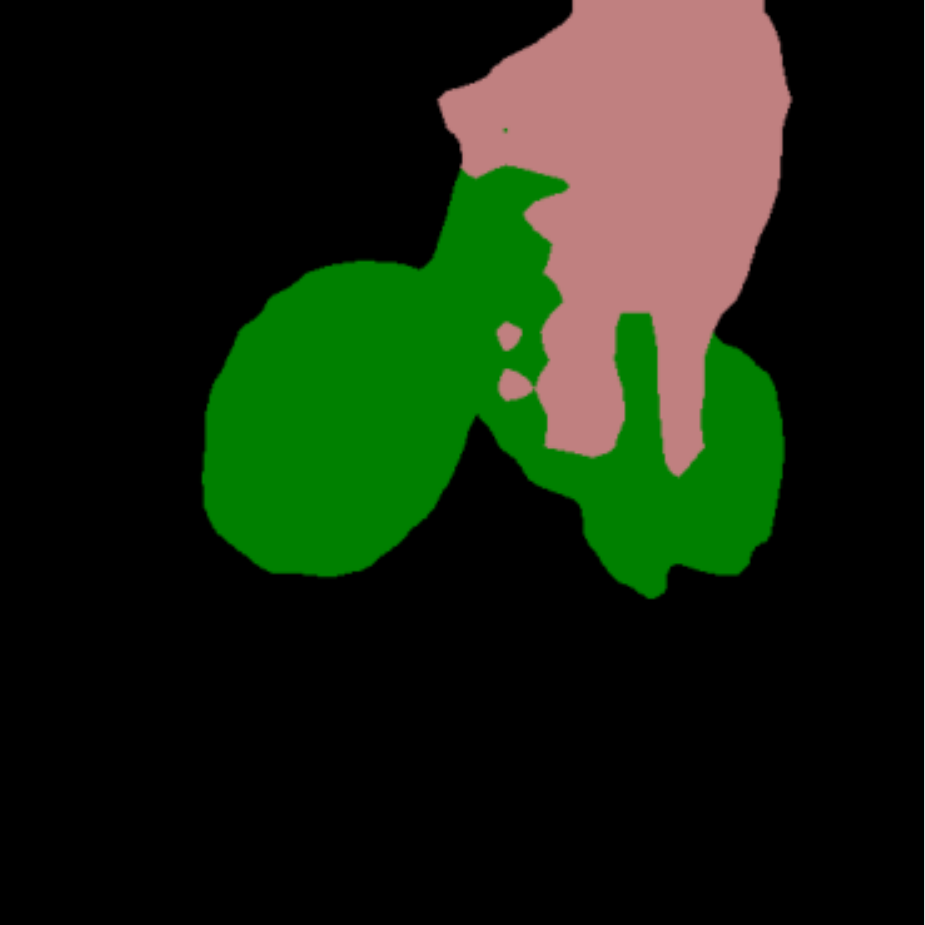}
    }
    \subfloat[\label{fig:sup:d} MIB\cite{Cermelli_2020_CVPR}]{
    \includegraphics[width=0.15\linewidth]{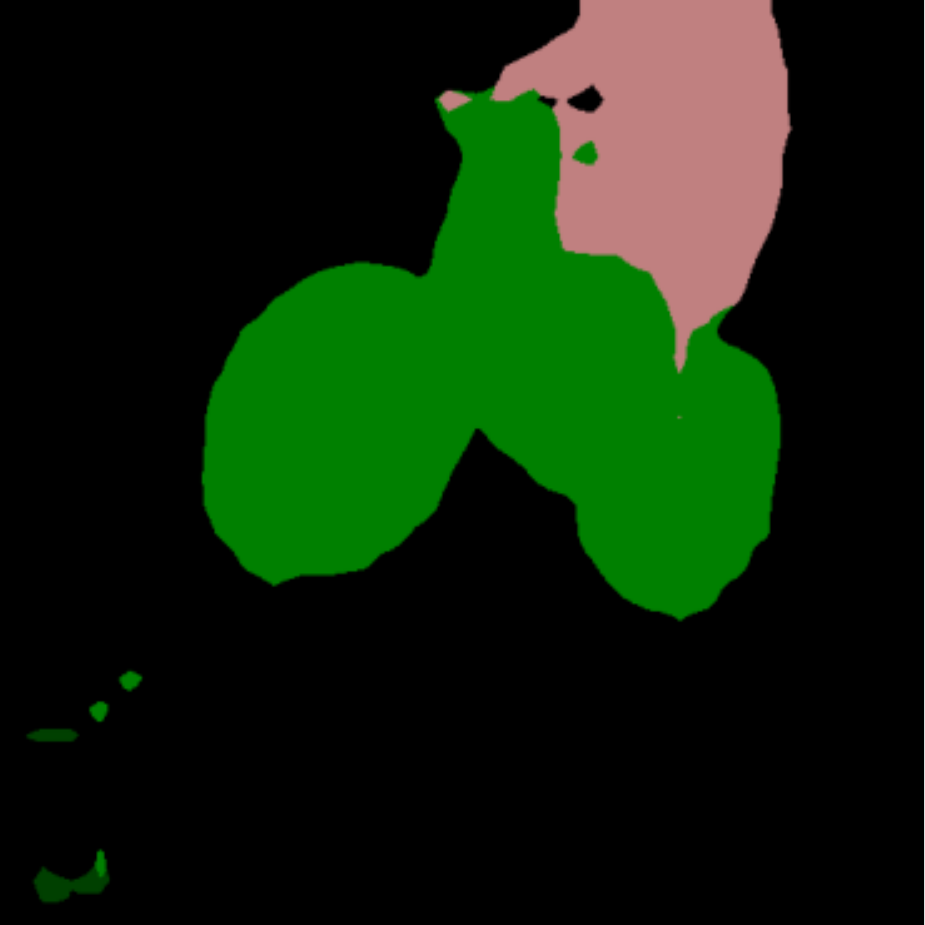}
    }
    \subfloat[\label{fig:sup:e} PLOP\cite{douillard2021plop}]{
    \includegraphics[width=0.15\linewidth]{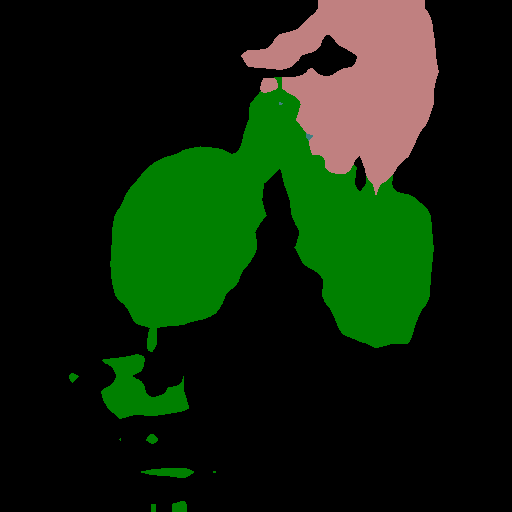}
    }
    \subfloat[\label{fig:sup:f} Ours]{
    \includegraphics[width=0.15\linewidth]{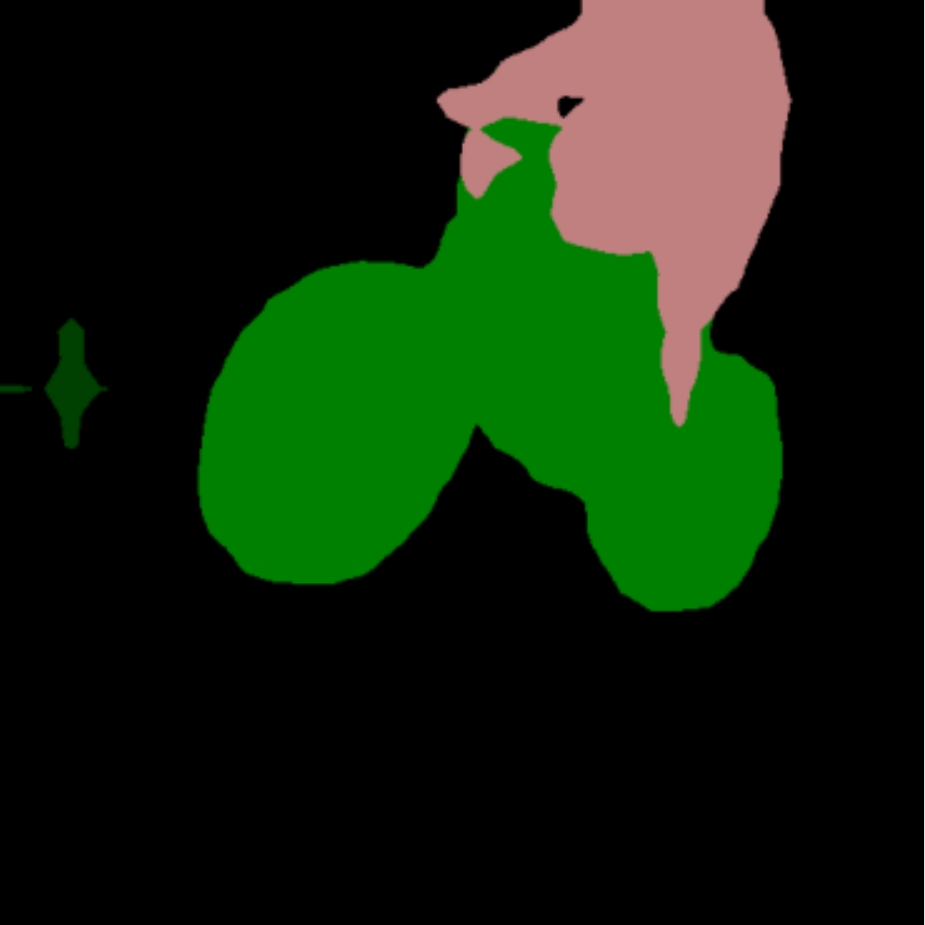}
    } \\
    \subfloat[\label{fig:sup:g} RGB]{
    \includegraphics[width=0.15\linewidth]{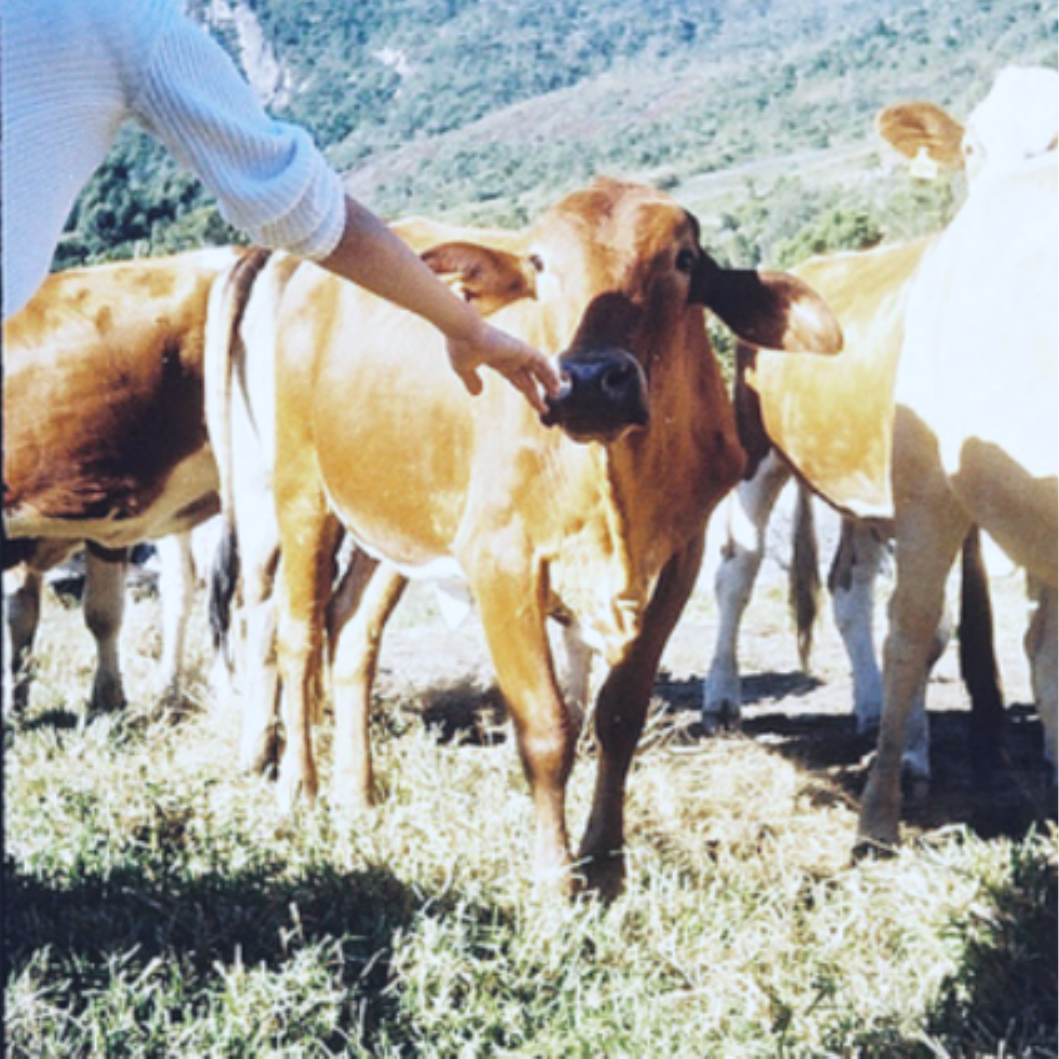}
    }
    \subfloat[\label{fig:sup:h} GT]{
    \includegraphics[width=0.15\linewidth]{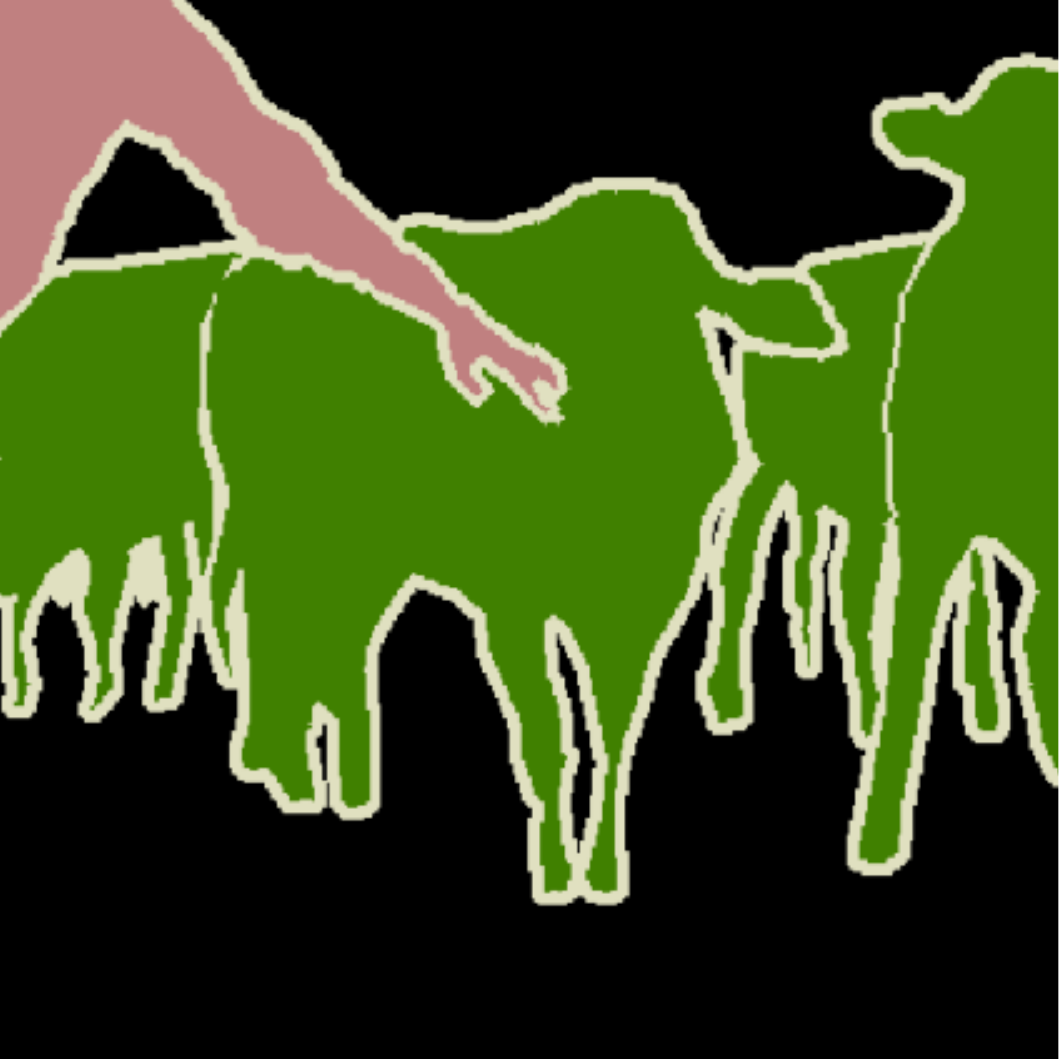}
    }
    \subfloat[\label{fig:sup:i} FT (ours)]{
    \includegraphics[width=0.15\linewidth]{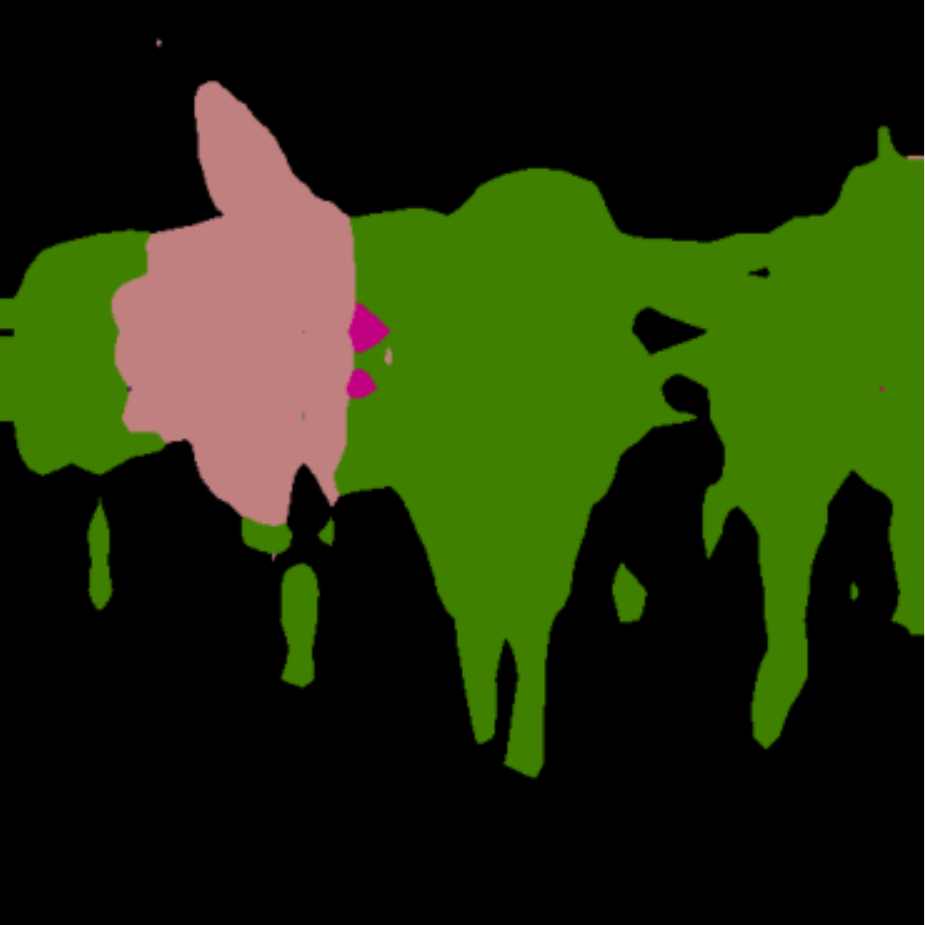}
    }
    \subfloat[\label{fig:sup:j} MIB\cite{Cermelli_2020_CVPR}]{
    \includegraphics[width=0.15\linewidth]{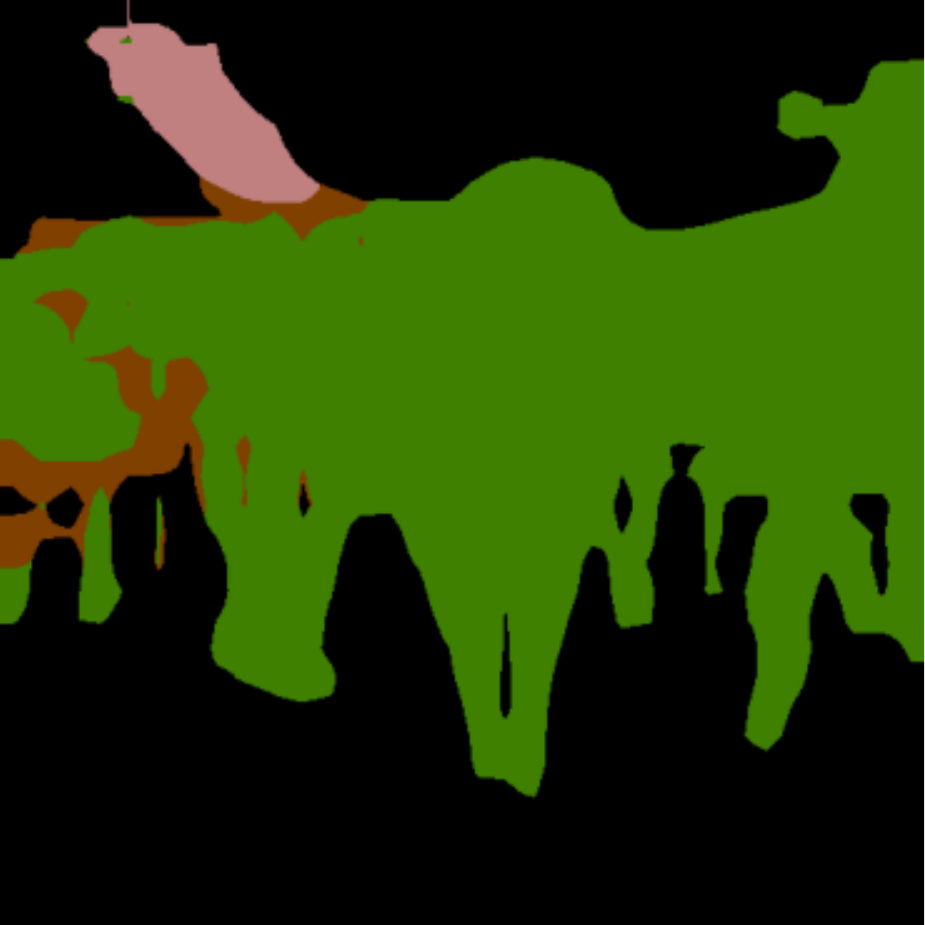}
    }
    \subfloat[\label{fig:sup:k} PLOP\cite{douillard2021plop}]{
    \includegraphics[width=0.15\linewidth]{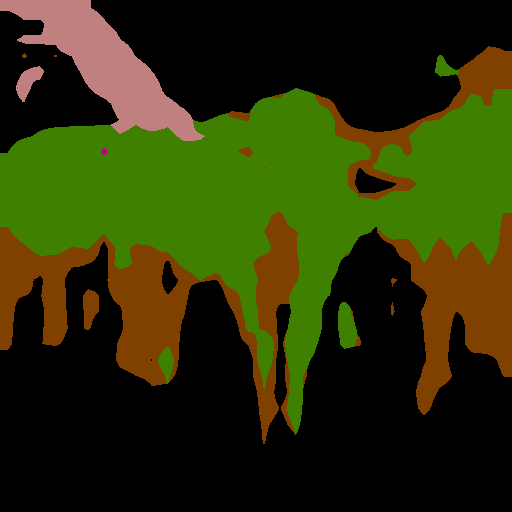}
    }
    \subfloat[\label{fig:sup:l} Ours]{
    \includegraphics[width=0.15\linewidth]{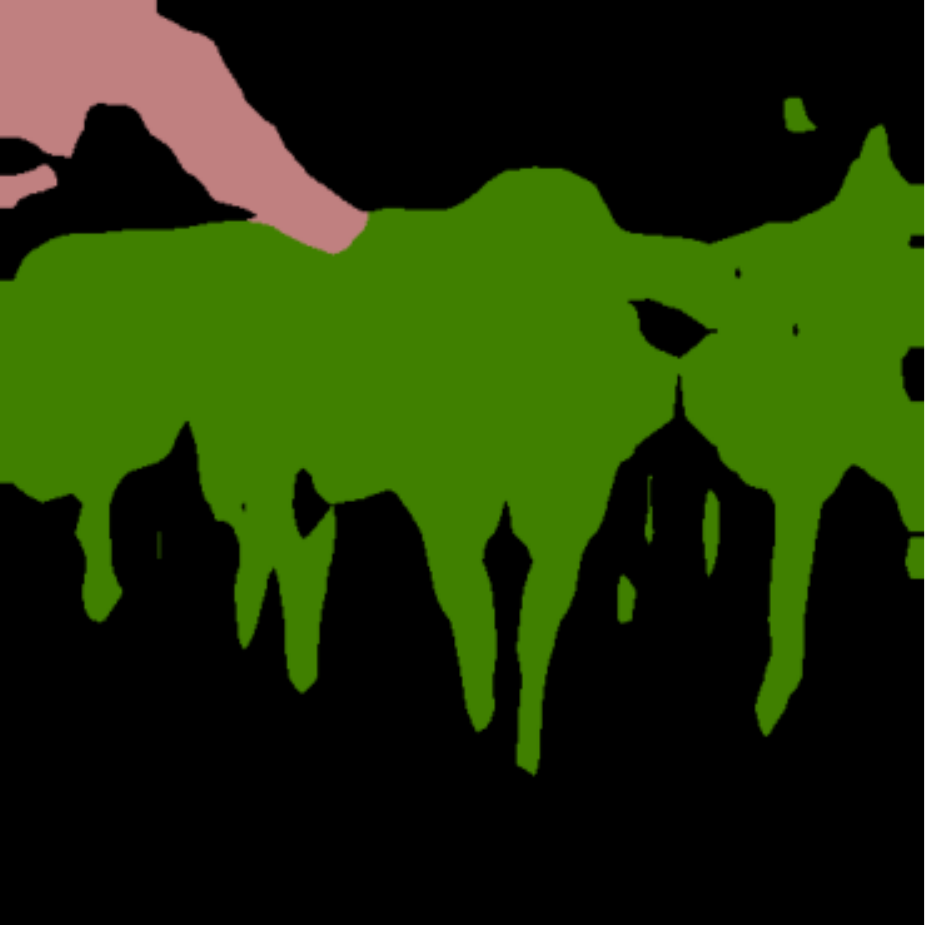}
    } \\
    \subfloat[\label{fig:sup:m} RGB]{
    \includegraphics[width=0.15\linewidth]{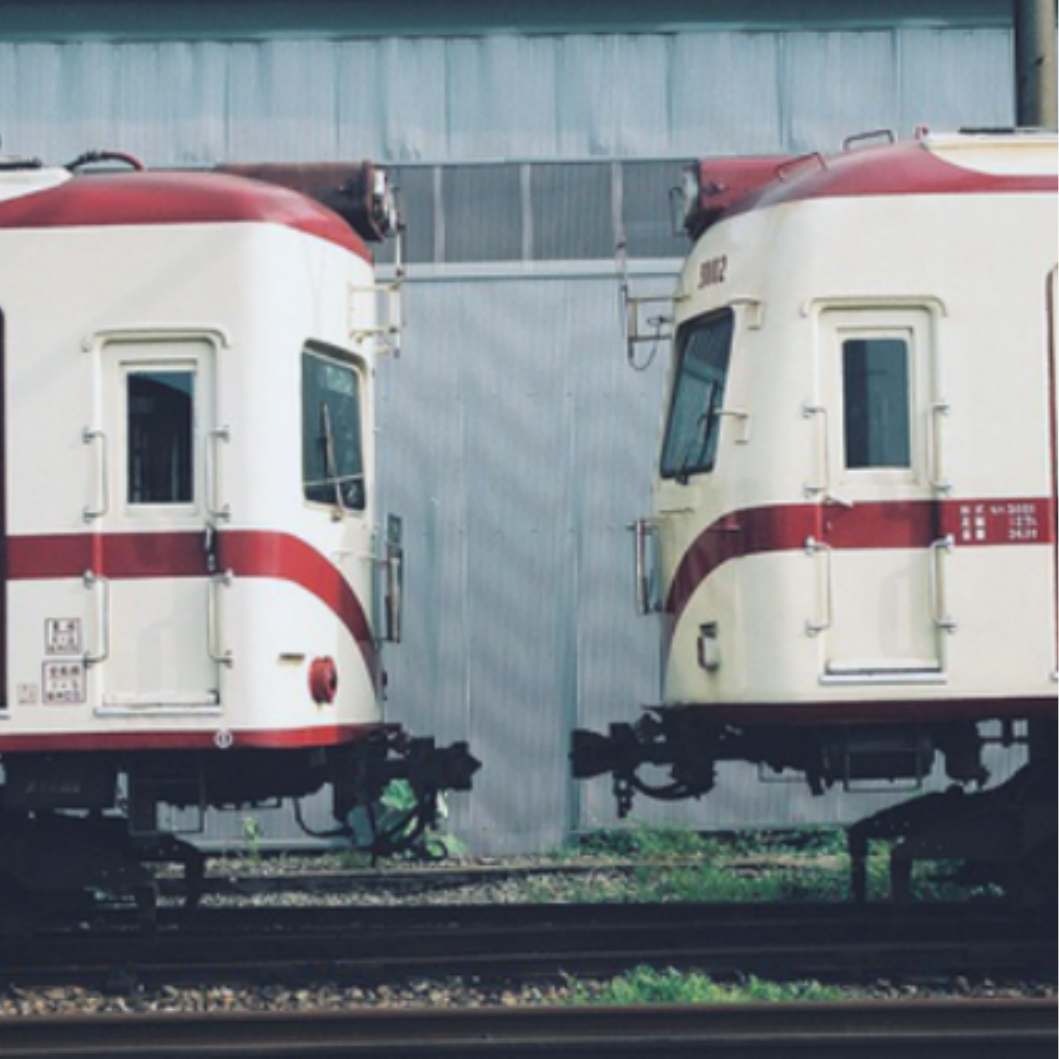}
    }
    \subfloat[\label{fig:sup:n} GT]{
    \includegraphics[width=0.15\linewidth]{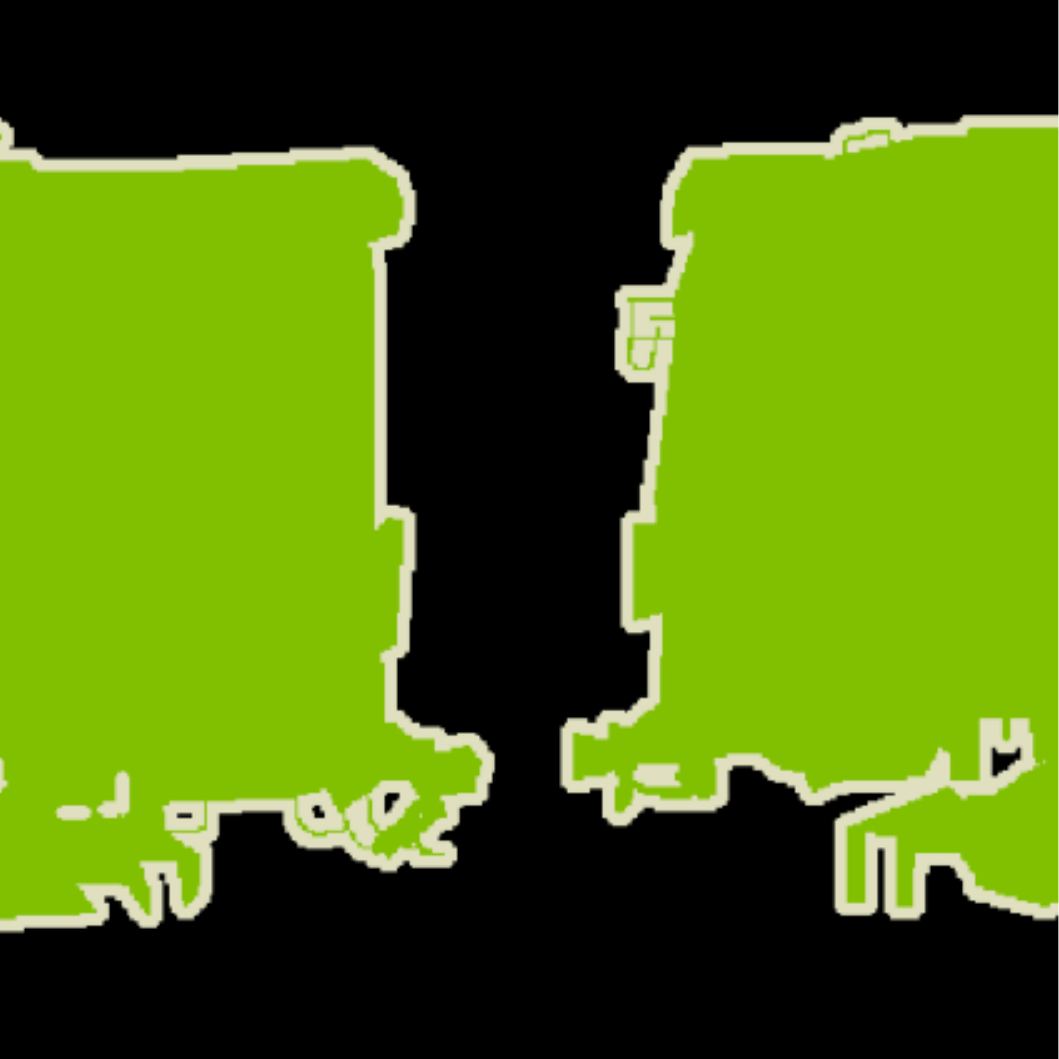}
    }
    \subfloat[\label{fig:sup:o} FT]{
    \includegraphics[width=0.15\linewidth]{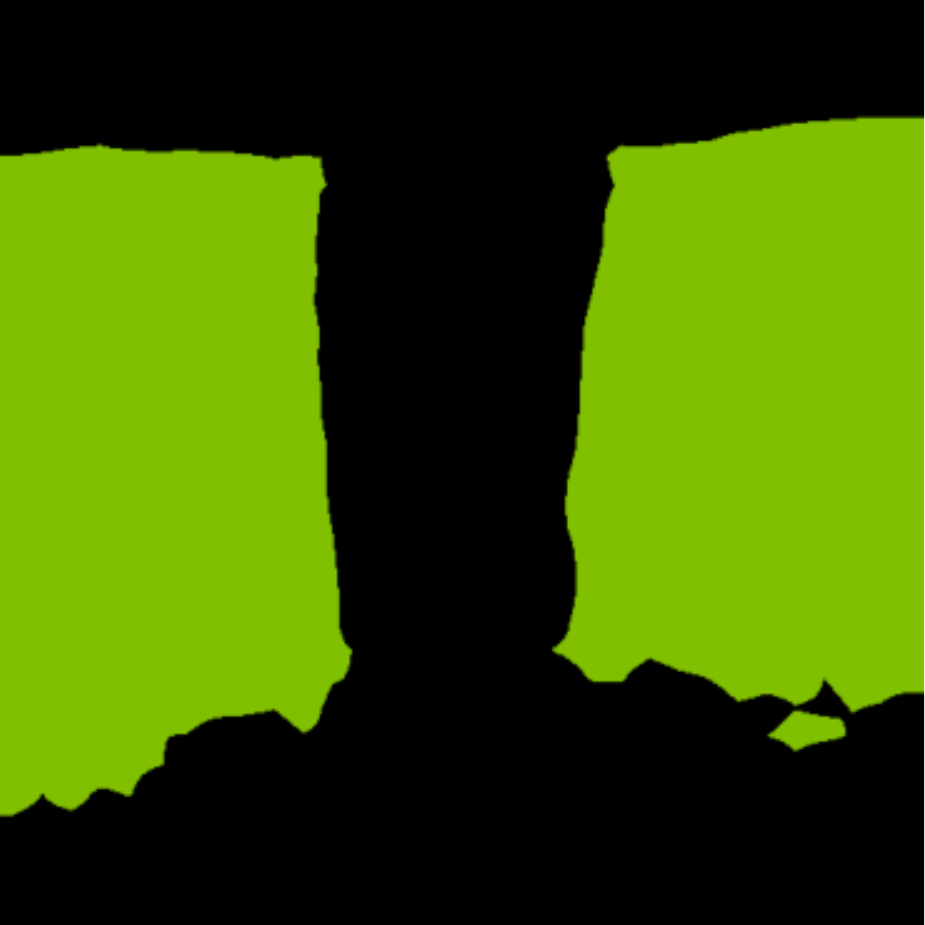}
    }
    \subfloat[\label{fig:sup:p} MIB\cite{Cermelli_2020_CVPR}]{
    \includegraphics[width=0.15\linewidth]{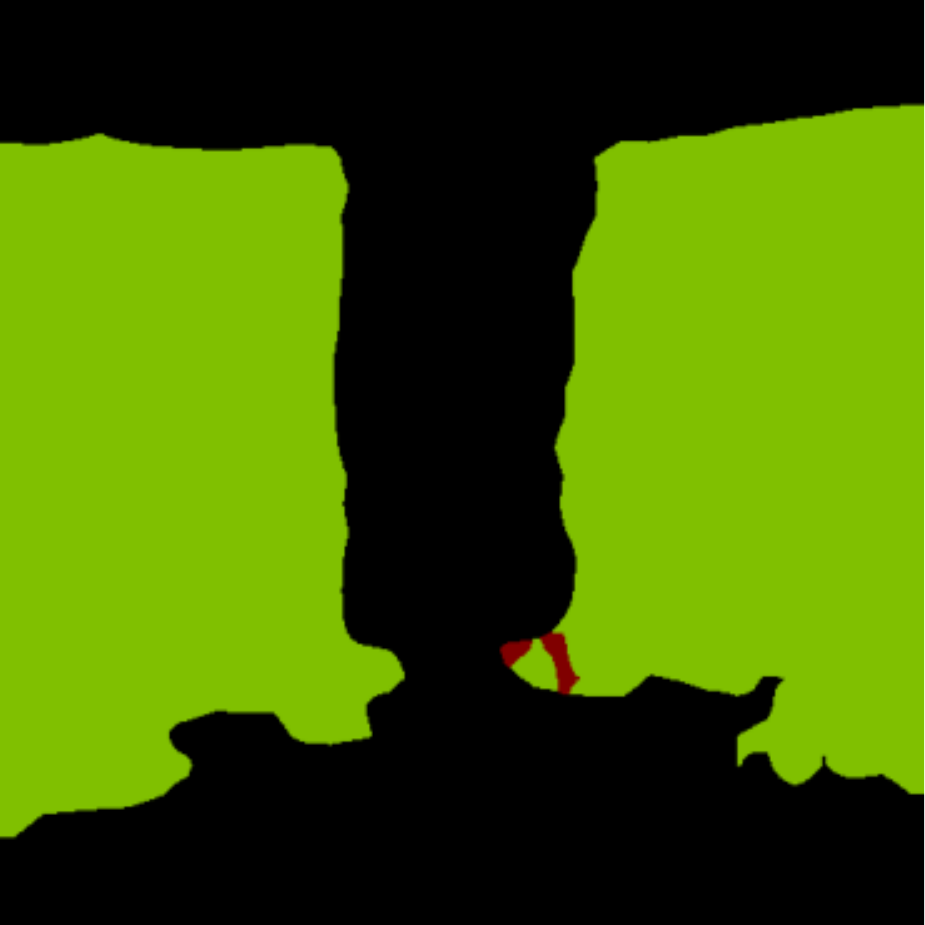}
    }
    \subfloat[\label{fig:sup:q} PLOP\cite{douillard2021plop}]{
    \includegraphics[width=0.15\linewidth]{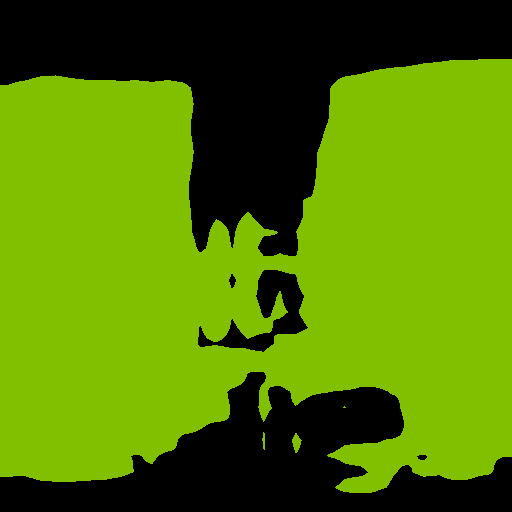}
    }
    \subfloat[\label{fig:sup:r} Ours]{
    \includegraphics[width=0.15\linewidth]{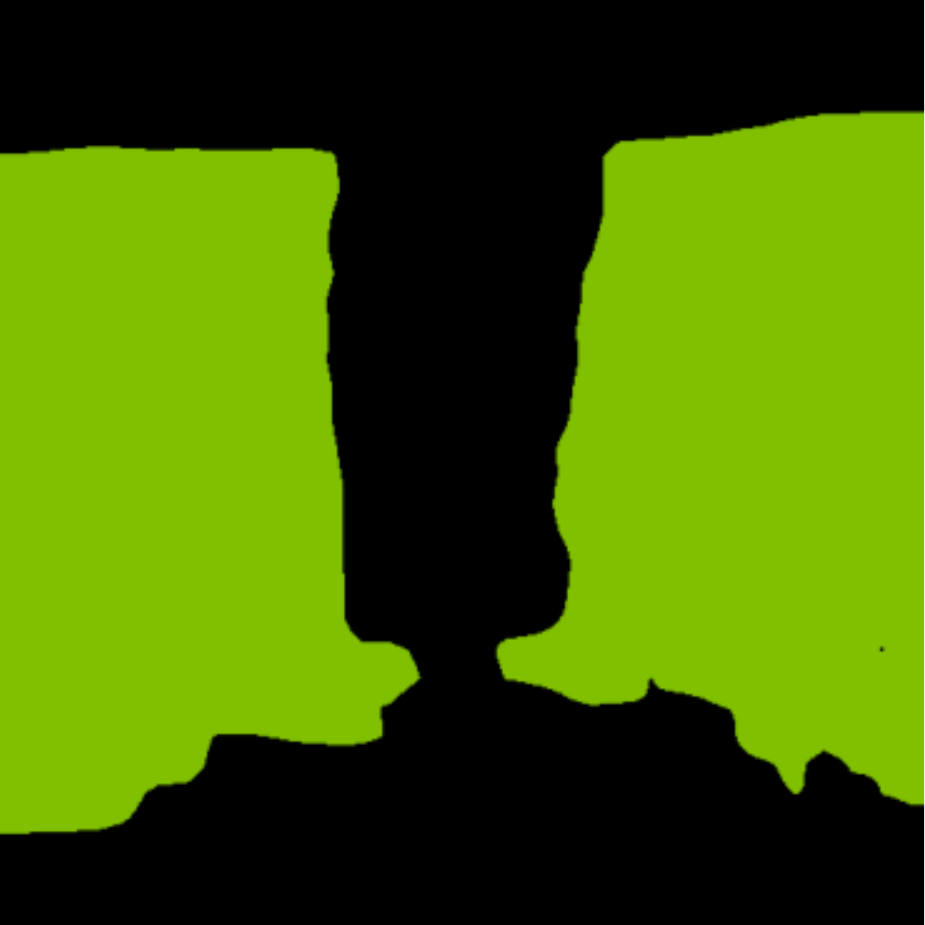}
    }
    \caption{Qualitative example on Pascal VOC 15-5.
    }
    \label{fig:sup:15-5:ov}
\end{figure*}

This section provides some general qualitative results on the Pascal VOC \cite{pascal-voc-2012} \textit{15-5} \textit{overlap} setting in fig. \ref{fig:sup:15-5:ov}. 
They illustrating that our method manages finer details better and avoids spurious classifications in the background. The results from fine-tuning are surprisingly good when using our proposed implicit background modeling, showing a tendency of classes bleeding out but without introducing spurious classifications, see panel (c). This illustrates how the implicit background model facilitates maintaining previous knowledge of the model, even without any explicit constraints during learning.

\begin{figure*}
    \centering
    \subfloat[\label{fig:sup:1} RGB]{
    \includegraphics[width=0.19\linewidth]{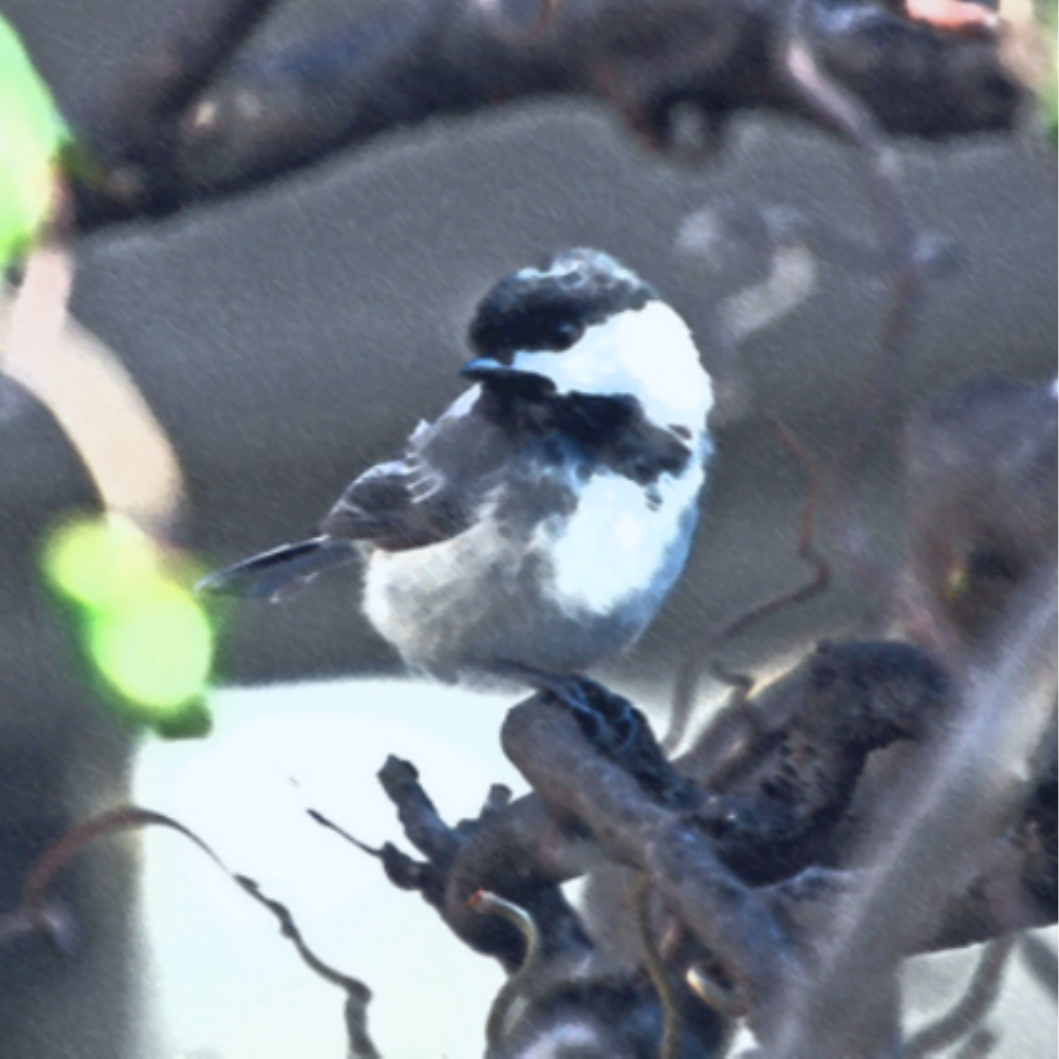}
    }
    \subfloat[\label{fig:sup:2} FT (15-5)]{
    \includegraphics[width=0.19\linewidth]{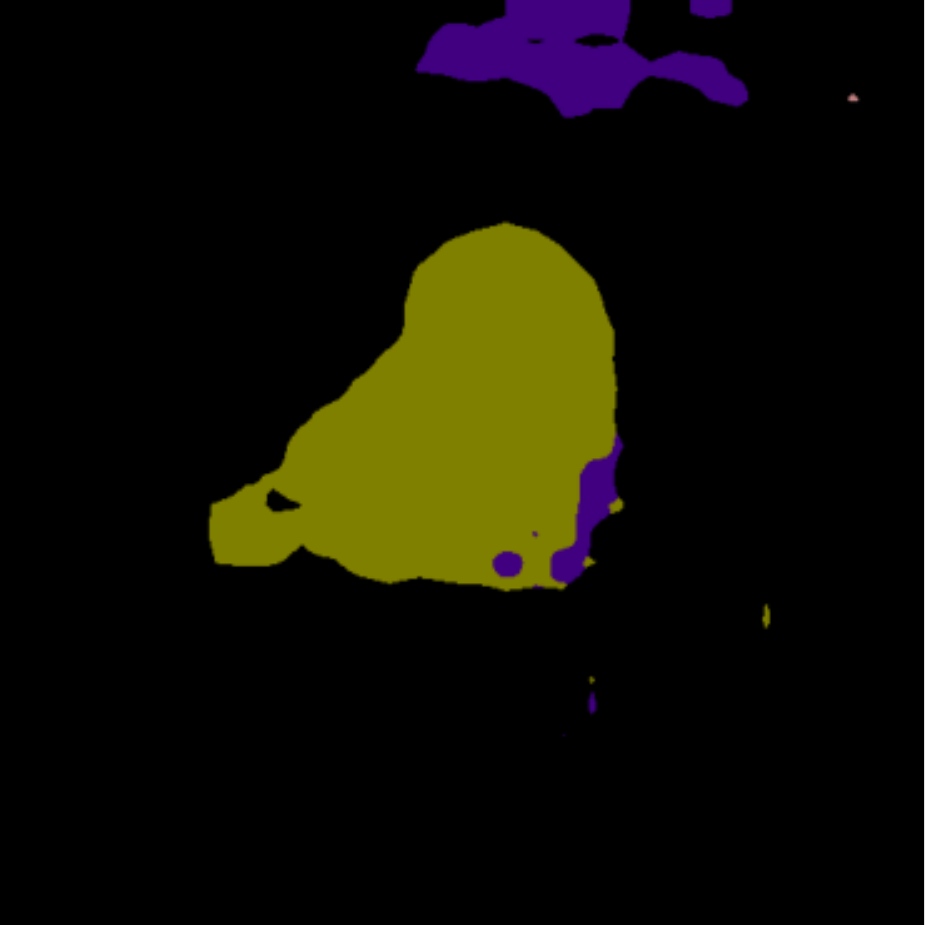}
    }
    \subfloat[\label{fig:sup:3} MIB (15-5)]{
    \includegraphics[width=0.19\linewidth]{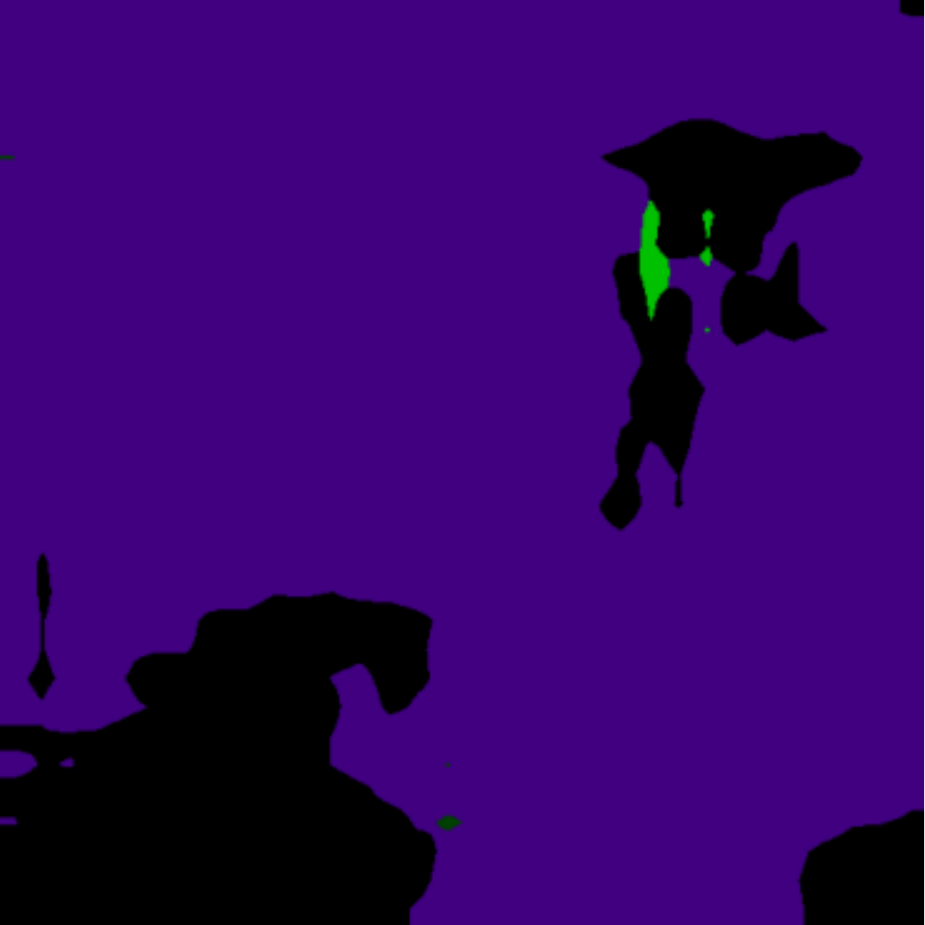}
    }
    \subfloat[\label{fig:sup:4} PLOP (15-5)]{
    \includegraphics[width=0.19\linewidth]{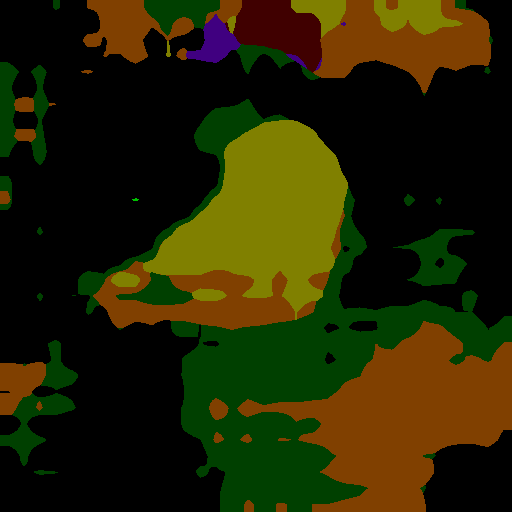}
    }
    \subfloat[\label{fig:sup:5} Ours (15-5)]{
    \includegraphics[width=0.19\linewidth]{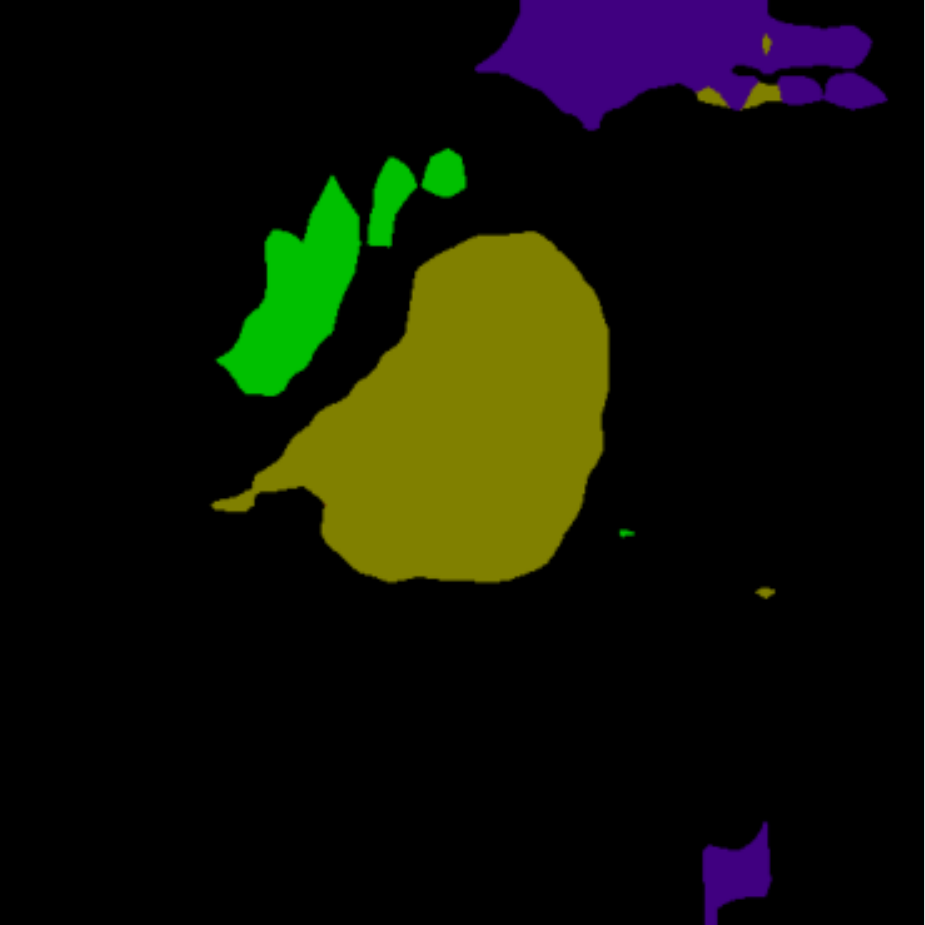}
    }
    \\
    \subfloat[\label{fig:sup:6} GT]{
    \includegraphics[width=0.19\linewidth]{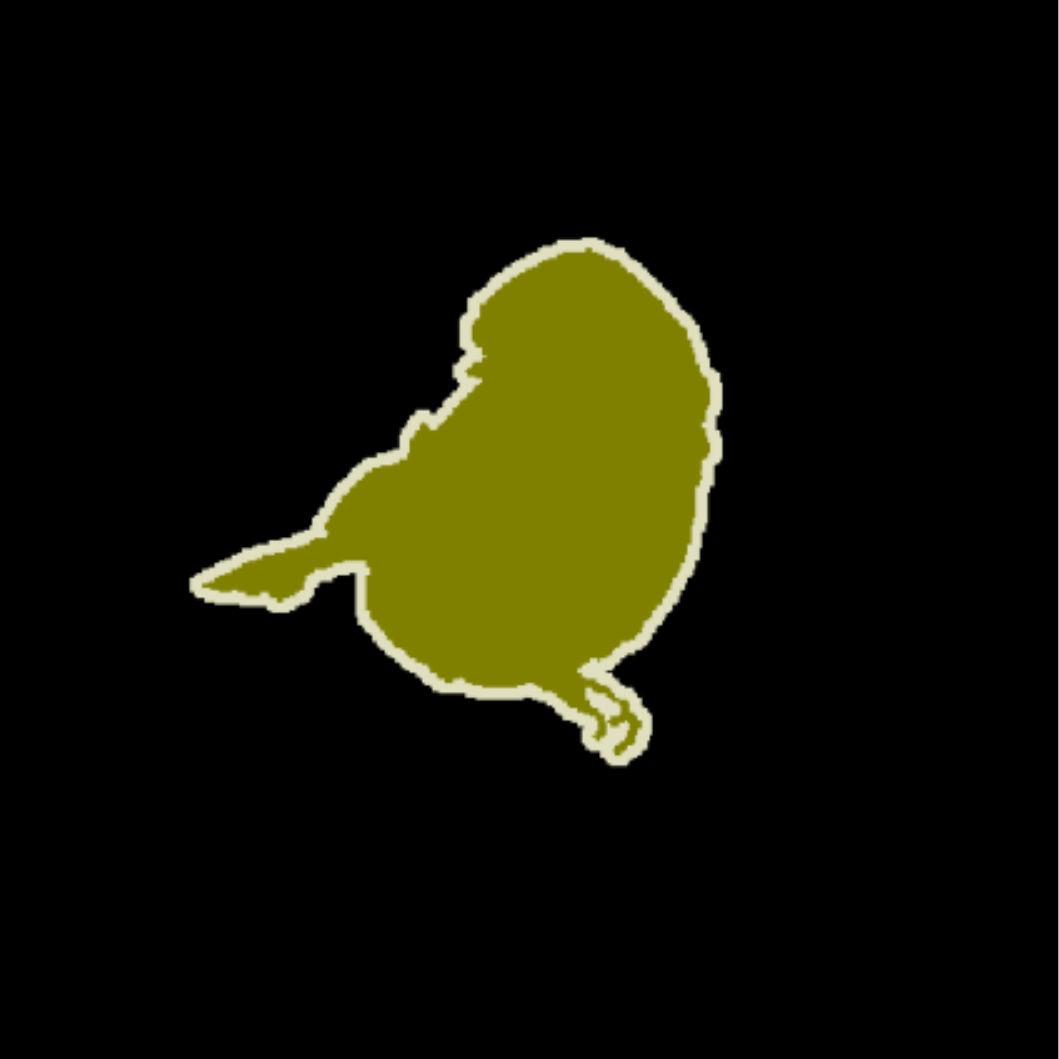}
    }
    \subfloat[\label{fig:sup:7} FT (15-1)]{
    \includegraphics[width=0.19\linewidth]{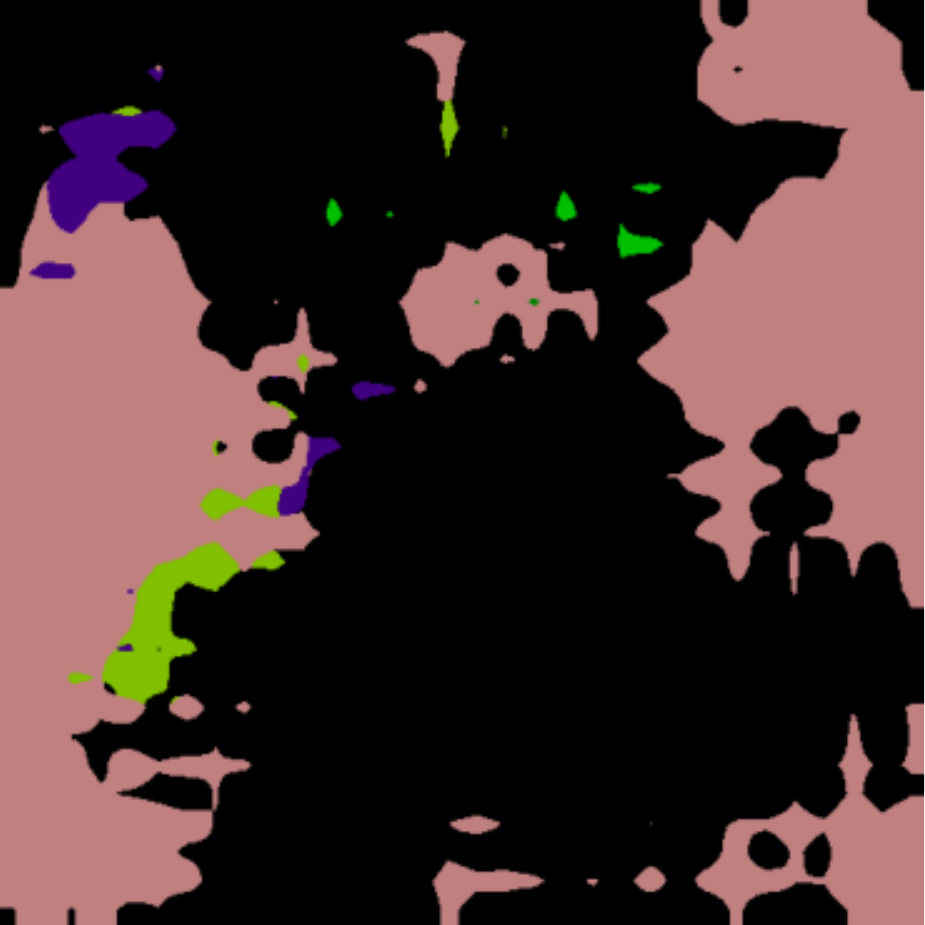}
    }
    \subfloat[\label{fig:sup:8} MIB (15-1)]{
    \includegraphics[width=0.19\linewidth]{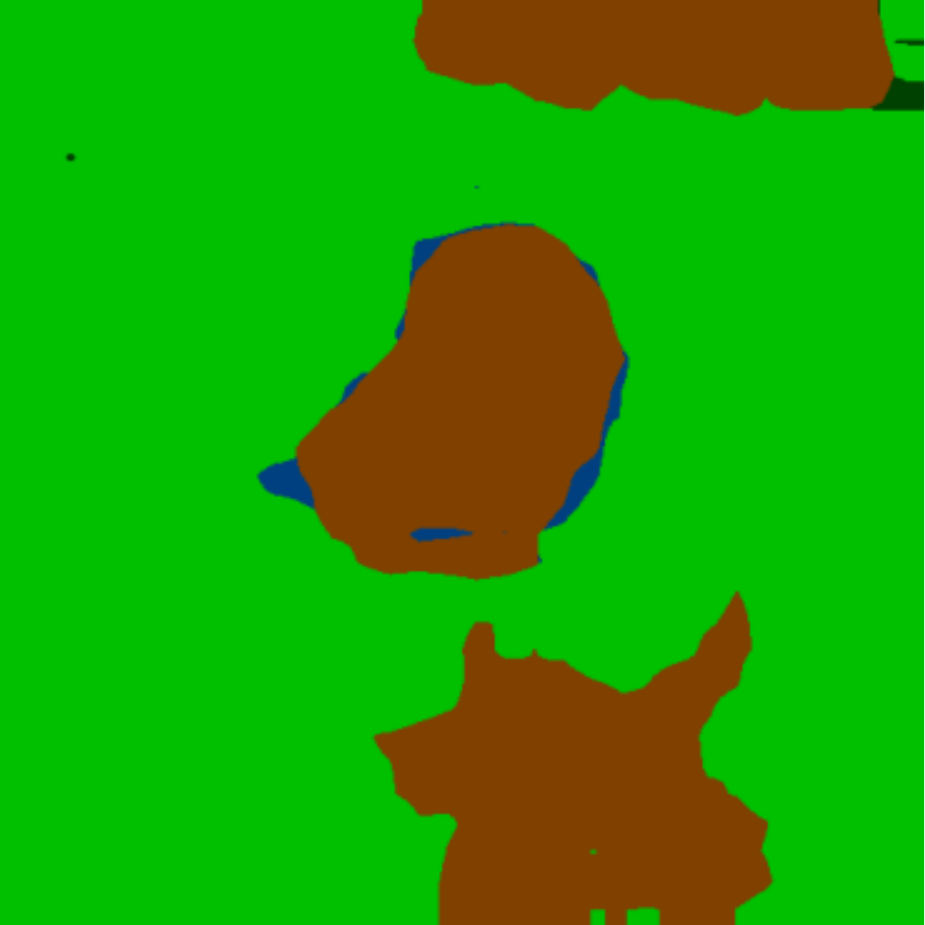}
    }
    \subfloat[\label{fig:sup:9} PLOP (15-1)]{
    \includegraphics[width=0.19\linewidth]{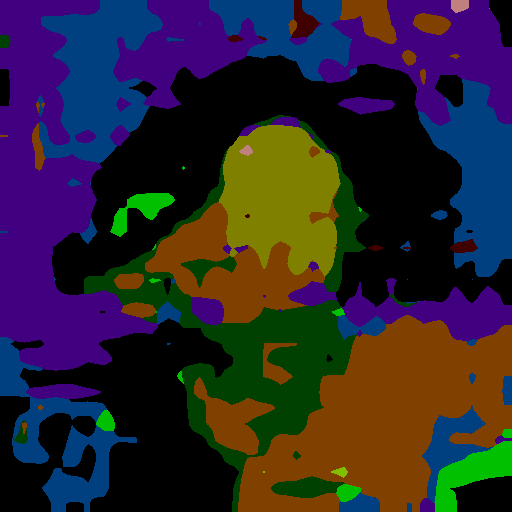}
    }
    \subfloat[\label{fig:sup:0} Ours (15-1)]{
    \includegraphics[width=0.19\linewidth]{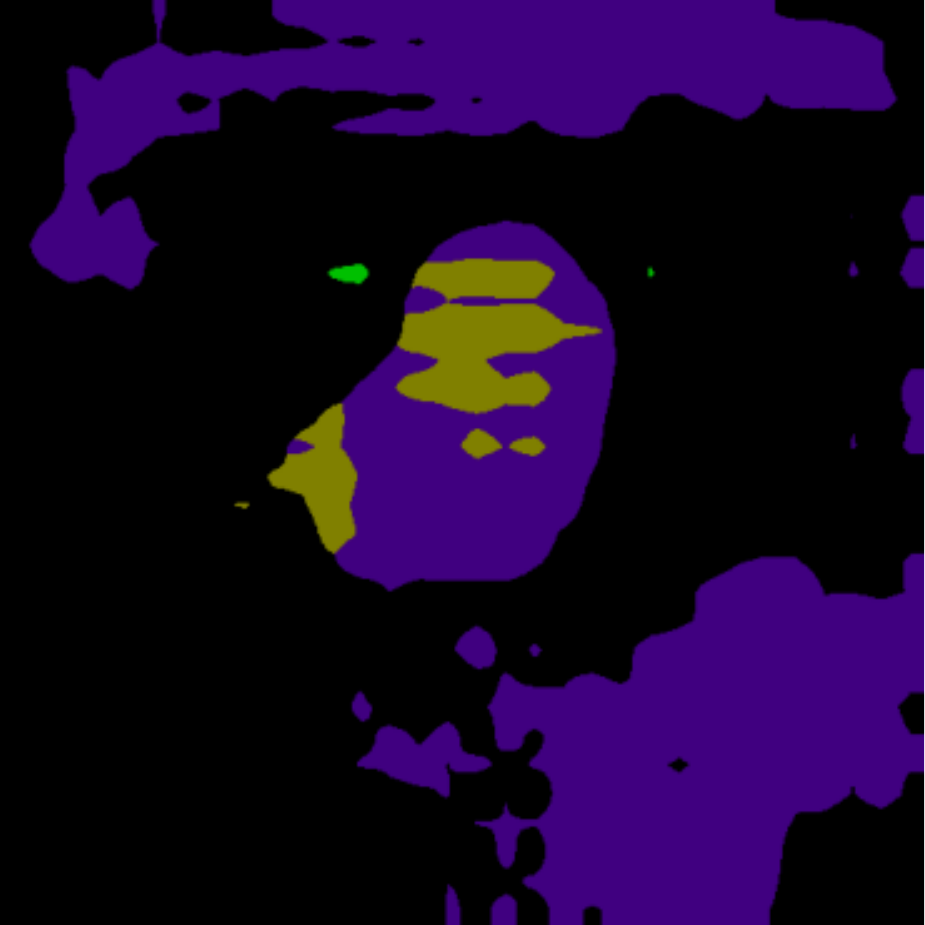}
    }
    \caption{Qualitative example (failure case) on Pascal VOC, comparing prediction on the \textit{15-1} and \textit{15-5} tasks
    }
    \label{fig:sup:15_5vs15_1}
\end{figure*}

Additionally, fig. \ref{fig:sup:15_5vs15_1} shows how the predictions can differ when learning multiple increments (\textit{15-1}) instead of a single one (\textit{15-5}).
These results show that the out-of-focus background is complex for all methods to handle, resulting in large spurious classifications in the background. While our implicit background model handles the \textit{15-5} scenario pretty well in comparison to the state-of-the-art, both our and other methods deliver drastically worse results on this failure case in the \textit{15-1} scenario.

\subsection{Results on ADE20k}
This section presents qualitative results on one scene from ADE20k \cite{zhou2019semantic} in order to illustrate the impact that different class orders have on the final predictions and how the foreground-background class-balancing can mitigate some of these problems.

ADE20k contains 150 different classes, ordered by frequency. The first set of classes, the base set, is composed of the first 50 or 100 classes depending on the task. Since these classes are the most frequent ones, the original order, the a-split, has a larger interclass imbalance between the base set and the following increments.
The b-split, however, is based on a random reordering of the classes, done by Cermelli et.al. \cite{Cermelli_2020_CVPR}, leading to a larger intraclass frequency variation but lower inter-class imbalance.

\begin{figure*}
    \centering
    \subfloat{
    \parbox[t][1cm][c]{0.25\linewidth}{\centering A-split: \\original class-order}
    }
    \subfloat{
    \hspace{1cm}
    }
    \subfloat{
    \parbox[t][1cm][c]{0.25\linewidth}{\centering B-split: \\ random class-order}
    }
    \\
    \centering
    \subfloat[\label{fig:sup:01} GT labels for a-split]{
    \includegraphics[width=0.3\linewidth]{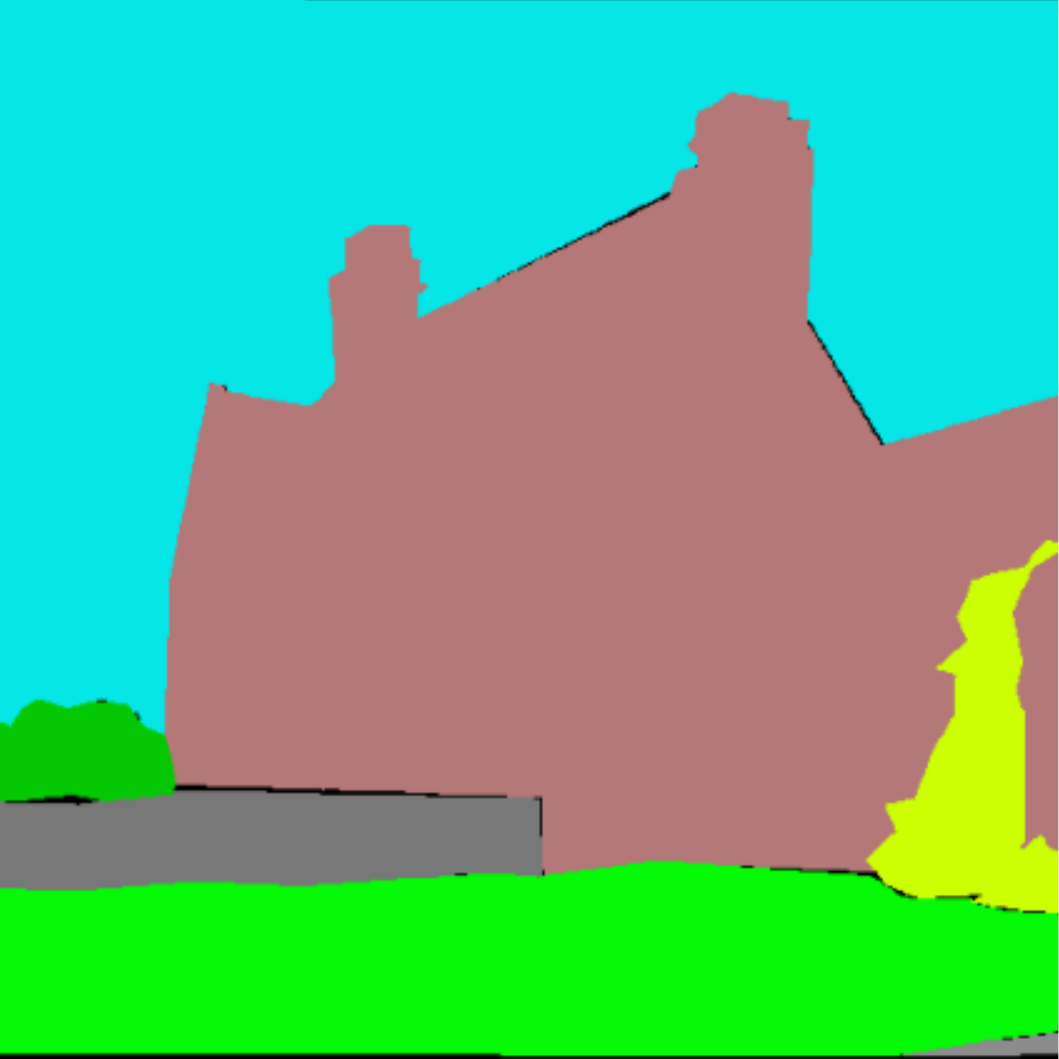}
    }
    \subfloat[\label{fig:sup:+2} RGB image]{
    \includegraphics[width=0.3\linewidth]{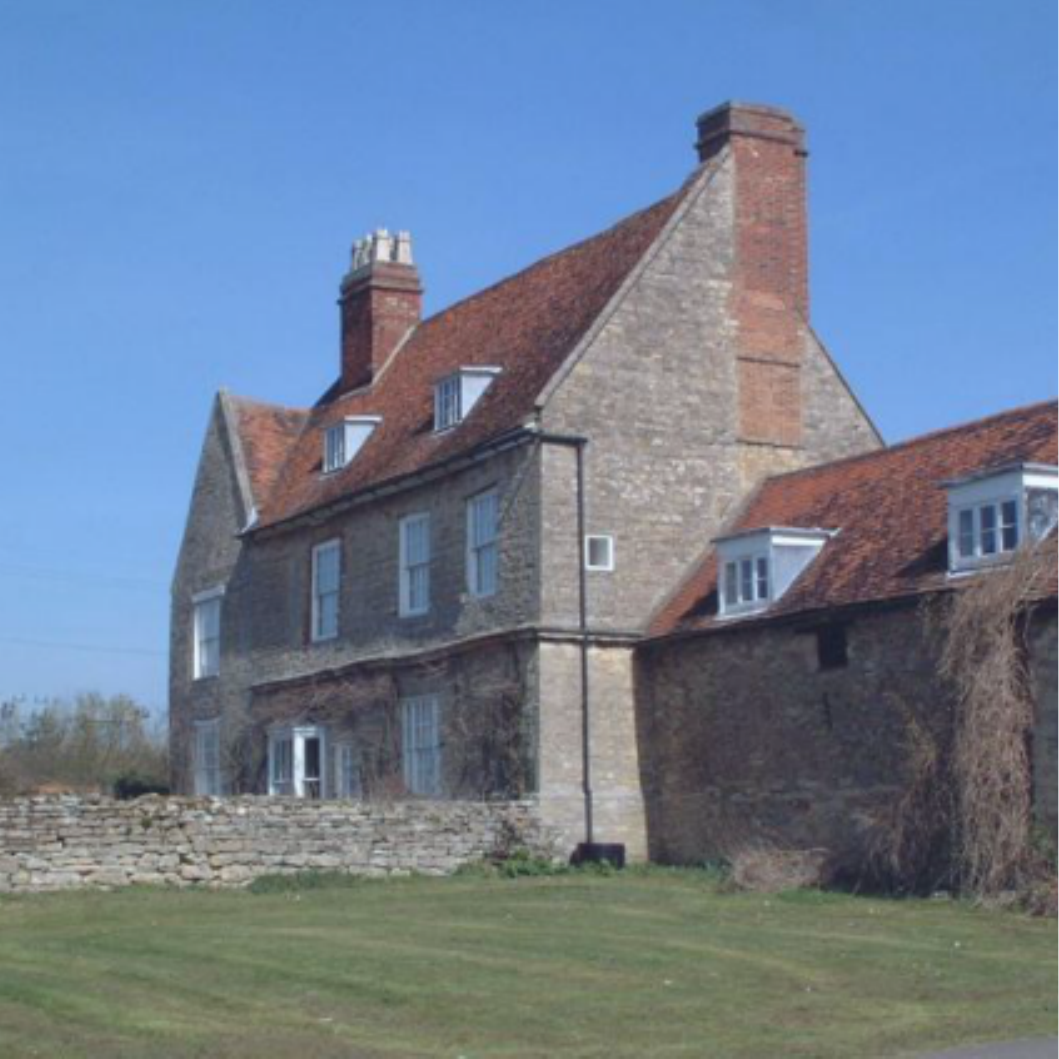}
    }
    \subfloat[\label{fig:sup:03} GT labels for b-split]{
    \includegraphics[width=0.3\linewidth]{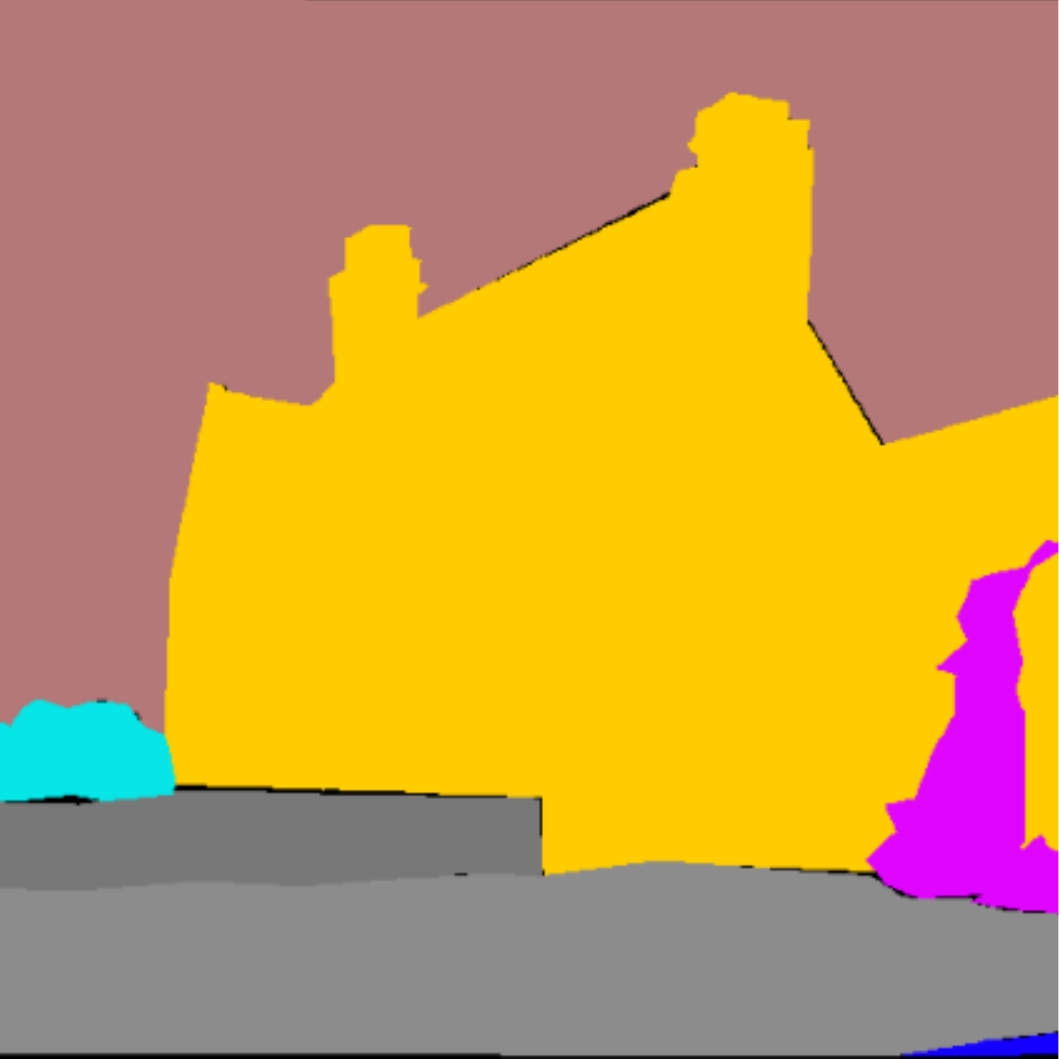}
    }
    \\
    \centering
    \subfloat[\label{fig:sup:04} w/o $w_\text{fg}$]{
    \includegraphics[width=0.22\linewidth]{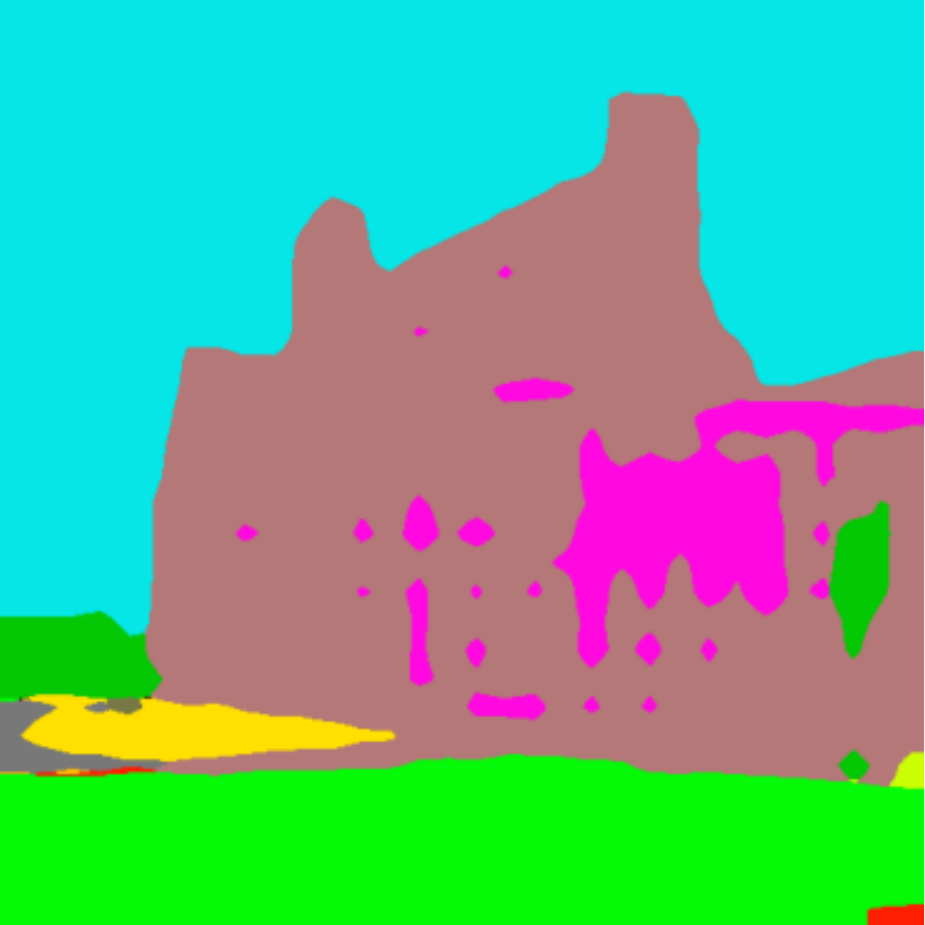}
    }
    \subfloat[\label{fig:sup:05} $w_\text{fg}$]{
    \includegraphics[width=0.22\linewidth]{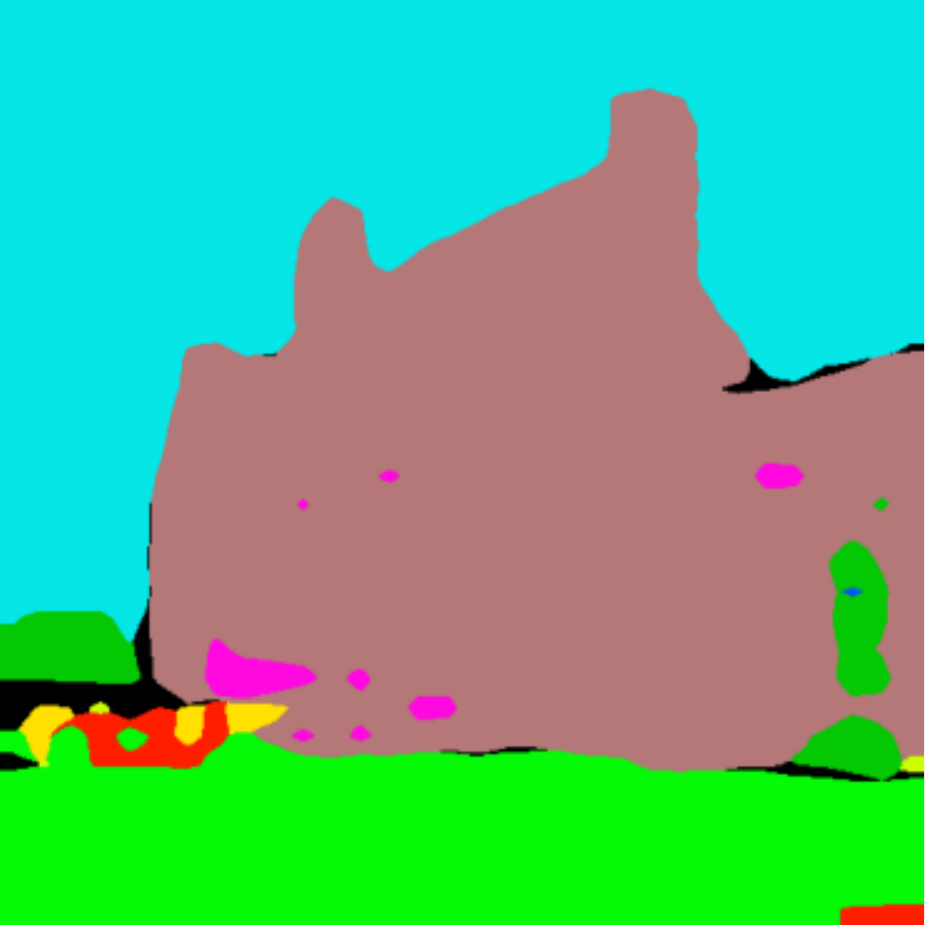}
    }
    \subfloat[\label{fig:sup:06} w/o $w_\text{fg}$]{
    \includegraphics[width=0.22\linewidth]{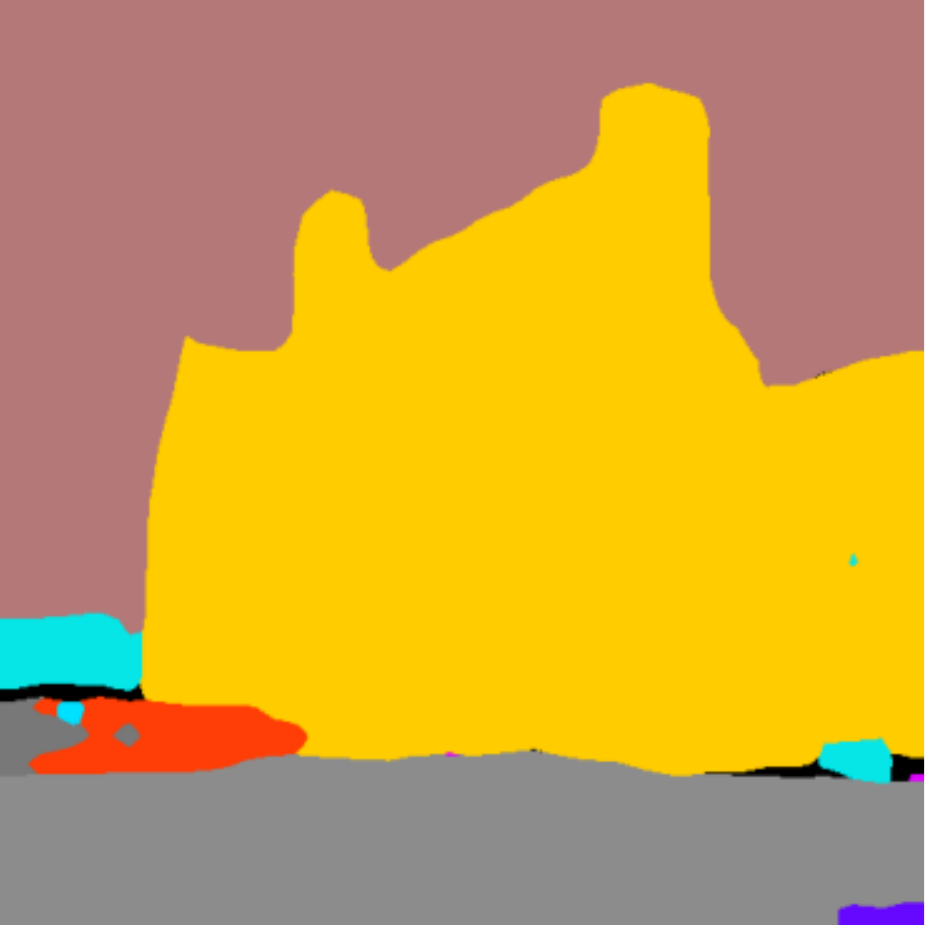}
    }
    \subfloat[\label{fig:sup:07} $w_\text{fg}$]{
    \includegraphics[width=0.22\linewidth]{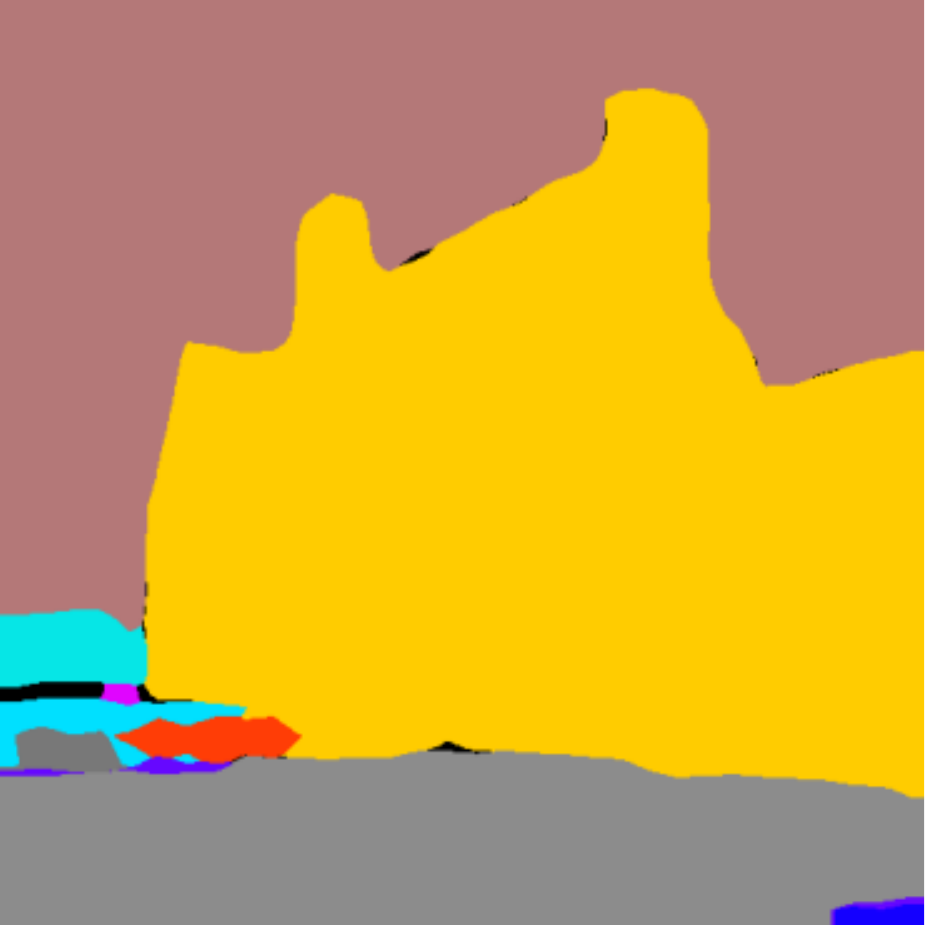}
    }
    \caption{Figure illustrating the effect that the class order and the foreground-background-balancing weight, $w_\text{fg}$, have on ADE20k \textit{100-10}.
    The a- and b-splits differ in the order in which classes are learned (a is the original order and b is an alternative order with a more balanced intra-increment class frequency).
    }
    \label{fig:sup:ade20k:building}
\end{figure*}

From fig. \ref{fig:sup:ade20k:building} it can be seen that the original class order lead to a larger confusion between the building and house classes unless the background foreground class balancing term is utilized. However, this confusion does not seem to occur for the b-split.

\clearpage

\end{document}